\documentclass[11pt]{article}
\usepackage{acl}
\usepackage{times}
\usepackage{latexsym}
\usepackage[T1]{fontenc}
\usepackage[utf8]{inputenc}
\usepackage{inconsolata}
\usepackage{graphicx}
\usepackage{booktabs}
\usepackage{multirow}
\usepackage{array}
\usepackage{amsmath}

\title{When Tool-Backed Skill Retrieval Fails:\\ Source-Style Collapse in Executable Capability Retrieval}
\author{
 \textbf{Yiqi Liu\textsuperscript{1}},\hspace{0.15cm}  
 \textbf{Joseph James\textsuperscript{2}}, \hspace{0.07cm} 
 \textbf{Yang Wang\textsuperscript{1}} \\ 
 \textbf{Chenghao Xiao\textsuperscript{3}},  \hspace{0.07cm}
 \textbf{Chenghua Lin\textsuperscript{1}}\vspace{0.2cm}
\\
 \textsuperscript{1}University of Manchester, \hspace{0.07cm}  
  \textsuperscript{2}University of Sheffield,  \hspace{0.07cm} \\
 \textsuperscript{3}Shanghai University of Finance and Economics
\\ \vspace{1.2cm}
\hspace{.1cm} \texttt{yiqi.liu@manchester.ac.uk} \hspace{.1cm} 
\texttt{chenghua.lin@manchester.ac.uk}
\\
}

\begin{document}
\maketitle
\begin{abstract}
Large-scale agents increasingly rely on retrieval to access external capabilities. We study this retrieval gate in structured tools and APIs, a measurable class of tool-backed executable skills that must be surfaced before an agent can plan, incorporate, or act. In this setting the retrieval layer can silently fail even when the capability corpus is fixed: on ToolRet, a retriever fine-tuned on one source-specific slice collapses on another source-specific slice of the \emph{same} benchmark, with \texttt{FT-1100} 
falling to \texttt{0.7\%} coverage on APIGen 
despite its higher lexical overlap with the gold tools. We call this failure mode \emph{source-style collapse}. Query-side TF-IDF fingerprints flag source styles on which the fine-tuned retriever is likely to fail better than semantic or length-based proxies, giving a cheap signal for mismatch over a fixed tool corpus. We propose ToolScout, a source-aware routing method that uses this signal as a routing guard: on the mixed \texttt{4,996}-query stream, TF-IDF-based routing raises coverage from \texttt{22.3\%} to \texttt{86.1\%}, and across five collapsed sources \texttt{20} matched examples raise the coverage-weighted global top-1 proxy from \texttt{1.3\%} to \texttt{53.9\%}. The same failure and routing behaviors persist when tools are rerendered as executable skill cards, which rules out raw API-schema format as the sole cause.

\end{abstract}

\begin{figure*}[t!]
\centering
\includegraphics[width=\textwidth]{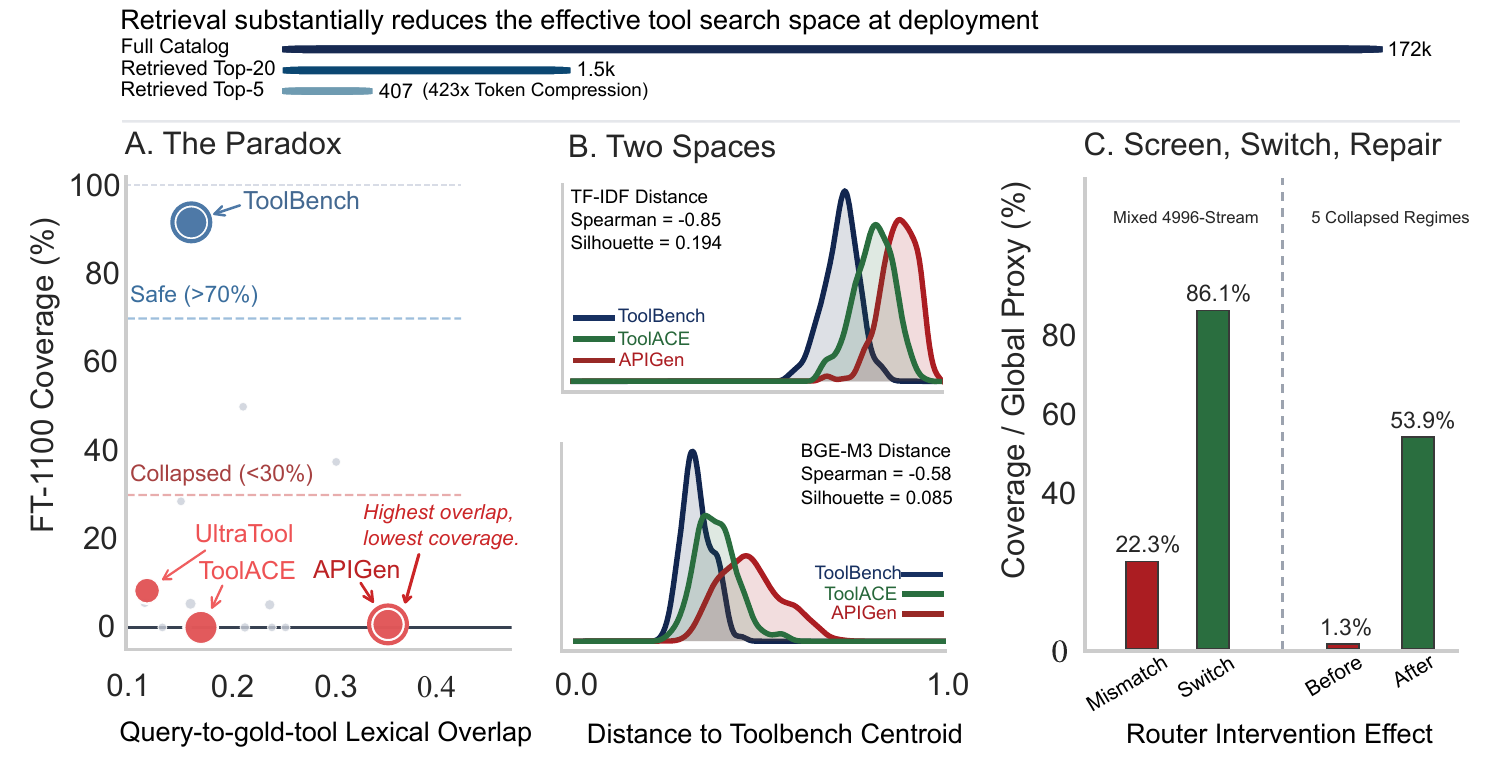}
\caption{Tool retrieval is source-style-sensitive on a fixed benchmark corpus. \textbf{Top:} retrieval compresses benchmark-native full-catalog exposure from about \texttt{172k} tokens to about \texttt{407} at retrieved top-\texttt{5}. \textbf{(A)} Lexical overlap fails to predict retrievability: \texttt{APIGen} has the highest query-to-gold-tool overlap yet collapses to \texttt{0.7\%} coverage under \texttt{FT-1100}. \textbf{(B)} TF-IDF distance separates source styles more clearly than \texttt{BGE-M3} semantic distance (\texttt{r=-0.85} vs.\ \texttt{-0.58}). \textbf{(C)} Route, switch, then repair: source-aware routing increases mixed-stream coverage by nearly fourfold, and \texttt{20} matched examples substantially improve the collapsed-source coverage-weighted global top-\texttt{1} proxy. }
\label{fig:toolscout-main}
\end{figure*}

\section{Introduction}

Agent systems are moving from short hand-crafted tool lists to large external capability pools. In this setting, retrieval decides which capabilities the model can see before reranking, planning, incorporation, or execution begins. We study the structured and executable case:  We render each tool or API as an executable skill card with a name, description, input schema, and implementation. This makes tool retrieval a controlled instance of executable capability retrieval, or Tool RAG. Similar source or style shifts are well known in document retrieval, while Tool RAG changes their operational meaning: when the right tool is missed, the agent loses an executable action instead of merely working with weaker evidence (If a travel-booking API never enters the candidate set, no planner can call it later).

We show that this entry layer can fail even when the tool corpus stays fixed. On ToolRet~\citep{huang2025toolret}, a checkpoint adapted on the \texttt{toolbench}-style \texttt{1100}-query slice remains strong on matching traffic but collapses on other source styles embedded in the same aggregated benchmark, including APIGen (\texttt{0.7\%} coverage), ToolACE (\texttt{0.0\%}), and UltraTool (\texttt{8.3\%}); see Appendix Table~\ref{tab:appendix-retriever-sources}. Lexical overlap alone does not explain the collapse: \texttt{APIGen} has higher lexical overlap with its gold tools than the training slice. Standard retrieval adaptation usually treats shift as a change in corpus, task, or broad query distribution. Here the benchmark-wide tool registry is fixed, and the failure is induced by the source style that generated the query-tool pairs. The stress case is therefore a same-corpus source-style shift.

This mismatch matters because realistic agent deployments rarely expose a small hand-crafted function list. They operate over hundreds or thousands of heterogeneous tools and APIs drawn from multiple providers, teams, and documentation styles~\citep{schick2023toolformer,patil2023gorilla,qin2024toolllm,li2023apibank,xu2024stabletoolbench,huang2025toolret}. In this pipeline, candidate generation first selects a small tool pool. Downstream rerankers or planners then operate on the selected candidates. If the gold tool never enters the pool, no downstream reranker or planner can recover it. Before any downstream planning or execution happens, the system must therefore decide which small subset of tools is worth exposing to the model. On StableToolBench, benchmark-native full-catalog exposure is approximately \texttt{172k} tokens, whereas dense top-\texttt{5} exposure is only about \texttt{407}. Tool retrieval is therefore what makes large-catalog prompting feasible at all. Once deployment depends on that compression, retrieval failures become systemic risks instead of local inefficiencies. The broader agent ecosystem only sharpens that pressure through multi-agent orchestration layers and tool-serving standards~\citep{wu2023autogen,li2023camel,hong2023metagpt,langchain_multiagent_docs,openai_agents_docs,anthropic_mcp_docs}.

 
 This paper studies that problem through \textbf{ToolScout}, a source-aware routing method for the retrieval gate. We ask whether a retriever adapted to one source-defined slice of an aggregated tool benchmark remains reliable on other source-defined slices when the tool corpus is unchanged. We use the term \emph{source-style collapse} to refer to the observed failure case in which the adapted retriever remains strong on matching-source queries but loses coverage on other source-defined slices over the same tools. We use \emph{source style} to mean the conventions introduced by an upstream query-generation or annotation process, including query wording, verbosity, schema-reference patterns, and the way queries are paired with tools in a fixed tool corpus. In simple terms, two sources may ask for the same tool in different dialects: one writes long ToolBench-style tasks, while another issues short API-template requests. The shift first becomes visible on the query side, because a retriever adapted on one source-defined slice can collapse on another over the same tools. Further analyses show that the target tool texts preserve a correlated, though weaker, fingerprint as well, so the practical query-side signal is one view of a broader paired shift.

Recent work on Skill Retrieval Augmentation studies a complementary downstream issue, showing that agents still need to incorporate retrieved skills effectively after relevant skills have been found~\citep{su2026sra}. ToolScout focuses on an earlier retrieval gate. When candidate generation misses the relevant tool, later reranking or planning stages have no correct tool to select. In this setting, the exposed capability is a structured tool or API and the tool corpus is fixed. We ask whether adapting a retriever to one source style can make it unreliable under source-style shift, and whether source-aware routing can recover the lost coverage.


A skill-card rerendering experiment confirms that raw API-schema formatting is an insufficient explanation: the same failure pattern, detector, and switch policy persist when tools are represented as executable skill cards.

ToolScout uses the query-side signal as a source-aware routing criterion. When a batch is likely to mismatch the source-specific retriever, ToolScout routes it to an aggregate-trained checkpoint and, when matched supervision becomes available, further adapts the retriever to recover the remaining coverage gap. This setting differs from reranking-focused work such as ToolRerank~\citep{wang2024toolrerank}, tool-representation work such as Tool2Vec~\citep{moon2024tool2vec}, and full agent stacks such as AnyTool~\citep{yang2024anytool} as ToolScout isolates the upstream routing layer itself under a shared retrieval and exposure protocol.

We make three linked contributions. First, using structured tools and APIs as measurable executable skill cards, we identify source-style collapse as a concrete retrieval failure mode: a dense retriever can fail on \texttt{APIGen}-style queries with \emph{higher} lexical overlap than the source it was trained on, so simple keyword access or query length fail to explain the performance decrease. Second, we characterize the conditions for failure. Across \texttt{19} source-query configurations, query-side TF-IDF distance is the strongest lightweight detector of source styles that are likely to mismatch the fine-tuned retriever (\texttt{78.9\%} leave-one-config-out accuracy), and the surrounding controls show that the signal is broader than a query-only rewrite effect. Third, we validate a source-aware routing method for this retrieval gate. On the mixed \texttt{4996}-query stream, TF-IDF-based routing raises coverage from \texttt{22.3\%} under a mismatched \texttt{FT-1100} retriever to \texttt{86.1\%}. Across five collapsed sources, \texttt{20} matched examples then raise the coverage-weighted global top-1 proxy from \texttt{1.3\%} to \texttt{53.9\%}. The routing criterion keeps source-specific retrievers on batches where they remain reliable and switches to the aggregate-trained checkpoint when source-style mismatch is likely to reduce coverage.

\section{Capability Retrieval and Tool Routing}

\paragraph{Skill and capability augmentation.}
Recent skill-centric work treats reusable skills as external capability units for agents~\citep{li2026skillsbench,zhou2026agenticskills,li2026singleagentskills}. Concurrent work on Skill Retrieval Augmentation (SRA) formulates scalable agent capability access as retrieving reusable skills from a large external skill corpus, and introduces SRA-Bench for decomposed evaluation of retrieval, incorporation, and end-task execution~\citep{su2026sra}. These works establish the broader setting: agents must retrieve external capabilities before deciding whether and how to incorporate them. ToolScout studies an earlier reliability question at that retrieval gate. Before a retrieved skill or tool can be used correctly, the relevant capability must enter the exposed candidate set.

\paragraph{Tool and API retrieval.}
Tools and APIs are the structured, executable end of the capability spectrum. Toolformer~\citep{schick2023toolformer} and Gorilla~\citep{patil2023gorilla} established the general tool-use setting. API-Bank~\citep{li2023apibank}, ToolBench / ToolLLM~\citep{qin2024toolllm}, StableToolBench~\citep{xu2024stabletoolbench}, ToolHop~\citep{li2025toolhop}, and ToolRet~\citep{huang2025toolret} further sharpened evaluation around large-scale retrieval and multi-hop tool use. Their schemas and gold-tool annotations make them a controlled substrate for measuring capability retrieval failures.

Tool2Vec~\citep{moon2024tool2vec} studies tool-specific representations, while ToolRerank~\citep{wang2024toolrerank} studies hierarchy-aware re-ranking and adaptive truncation once a candidate pool is already given. AnyTool~\citep{yang2024anytool} and DeepAgent~\citep{li2026deepagent} use hierarchical retrieval inside broader agent stacks. Agent frameworks such as ReAct~\citep{yao2023react}, AutoGen~\citep{wu2023autogen}, CAMEL~\citep{li2023camel}, and MetaGPT~\citep{hong2023metagpt}, together with orchestration stacks such as LangChain~\citep{langchain_multiagent_docs}, the OpenAI Agents SDK~\citep{openai_agents_docs}, and MCP tooling~\citep{anthropic_mcp_docs}, make small exposed schema sets a practical requirement before downstream planning.

\paragraph{Retrieval adaptation under capability shift.}
Our setup builds on standard retrieval practice: DPR~\citep{karpukhin2020dpr}, ANCE~\citep{xiong2020ance}, ColBERT~\citep{khattab2020colbert}, and BERT re-ranking~\citep{nogueira2019bertrerank} established core dense and re-ranking baselines, while BEIR~\citep{thakur2021beir} made cross-domain transfer a standard concern. Modern encoders such as BGE-M3~\citep{chen2024bgem3} and Qwen3-Embedding~\citep{zhang2025qwen3embedding} provide strong off-the-shelf backbones. The closest general-retrieval literature is \emph{domain adaptation for dense retrieval}. GPL, InPars, Promptagator, and AugTriever use pseudo-labeling or synthetic query generation to adapt retrievers to new target distributions~\citep{wang-etal-2022-gpl,bonifacio2022inpars,dai2022promptagator,meng2022augtriever}. Our few-shot repair stage is closest in spirit to that line. Our diagnostic setting keeps the capability corpus fixed while changing the source slice that generated the query-tool pairs. That change can still break a narrow dense specialist. Similar shifts can lower relevance in document retrieval, but executable capability retrieval makes the consequence harsher because a missed tool removes an action from the agent's candidate set instead of merely weakening its evidence base.

\section{ToolScout Routing Policy}

Given a user query \texttt{q} and tool library \texttt{T}, ToolScout returns a top-\(k\) candidate set that fits within the context-window token budget used to expose tools to the downstream model, while preserving enough coverage for downstream execution. In our main setting, this budget is operationalized by the number of retrieved tool cards and their rendered token length, rather than by monetary cost. The core system is a direct two-stage pipeline, \texttt{query $\rightarrow$ candidates $\rightarrow$ reranking}. We study staged retrieval and hierarchy only as optional extensions around the core path, and we return to them after the main collapse result.

We use \emph{source style} to mean the conventions introduced by an upstream query-generation or annotation process, including query wording, verbosity, schema-reference patterns, and the way queries are paired with tools in a fixed tool corpus. The definition covers both query surface form and the query-tool pairing conventions exposed to the retriever. On the query side, this appears in the distribution of user requests. On the tool side, it appears in the target tool texts that the retriever actually encodes. The main stress case is a same-corpus source-style shift: the tool registry stays fixed, while the phrasing, verbosity, and schema-reference habits of the query-tool pairs still change. When a 1100-trained checkpoint fails sharply under that shift, we refer to the resulting failure mode as \emph{source-style collapse}.

Formally, the direct candidate generator returns

\[
C(q)=\operatorname{top}_k(r(q,T)),
\]

where \(r\) is the lexical or dense retriever and \(k\) is the fixed candidate budget. If a query is decomposed into sub-queries \(\{q_1,\dots,q_m\}\), staged retrieval first scores tools for each sub-query and then keeps the top-\(k\) tools under the aggregated score:

{\footnotesize
\[
C_{\text{stage}}(q)
=
\operatorname{top}_k
\left(
\left\{
t \in T :
s(t)=\max_{j\in\{1,\dots,m\}} r(q_j,t)
\right\}
\right).
\]
}
Reranking then operates on the candidate set it actually receives.


Tools are rendered from structured metadata such as name, description, arguments, and tags, then encoded with a sentence-transformer backbone. We use \texttt{BAAI/BGE-M3}~\citep{chen2024bgem3} as the main dense backbone; Table~\ref{tab:main-offtheshelf} compares it with \texttt{bge-small-en-v1.5}, and Table~\ref{tab:appendix-retriever-ft-backbone-swaps} reports additional backbone swaps. Candidate generation uses either lexical or dense retrieval. The main dense pipeline retrieves the top-\texttt{k} tools with the encoder and applies a lightweight learned reranker over retrieval, lexical, and metadata features. Appendix Table~\ref{tab:appendix-reranker-stronger} reports stronger reranker variants, showing that improved reranking helps covered queries but does not remove the candidate-generation bottleneck.

Starting from an off-the-shelf sentence encoder, we fine-tune on query-tool pair supervision from ToolRet using MultipleNegativesRankingLoss (MNRL), the SentenceTransformers implementation of an in-batch contrastive objective closely related to InfoNCE~\citep{oord2018cpc}
with retrieval-confusing hard negatives mined from the current depth-\texttt{20} dense pool. This directly targets the bottleneck identified in the paper: if candidate generation is failing upstream, the cleanest repair is to improve the retriever before adding more downstream complexity.

The resulting routing policy is source-aware and can operate with multiple retriever checkpoints. For an incoming query set \(\mathcal{Q}\) and a retriever trained on source slice \(\mathcal{R}\), ToolScout computes a compatibility score
\[
d_{\text{tfidf}}(\mathcal{Q}, \mathcal{R}) = 1 - \cos\!\big(\mu_{\text{tfidf}}(\mathcal{Q}), \mu_{\text{tfidf}}(\mathcal{R})\big),
\]
where \(\mu_{\text{tfidf}}\) is the batch centroid in TF-IDF space. The safe/unsafe band for \(d_{\text{tfidf}}\) is calibrated once on the natural ToolRet source splits in leave-one-config-out form and then fixed for the routing experiments. If the resulting distance falls inside that known-safe band, direct routing with the matched fine-tuned retriever is the default path; if it falls in a clearly unsafe band, the system switches to a broader aggregate-trained checkpoint or keeps staged routing available while matched supervision is collected. Once a small amount of matched supervision becomes available, the same policy can trigger a short continuation fine-tuning step on the mismatched checkpoint. We evaluate this TF-IDF-based compatibility predictor in three settings: leave-one-configuration-out source mismatch detection, the main \texttt{1100} \(\rightarrow\) \texttt{4996} mixed-source transfer test, and few-shot matched adaptation on collapsed sources.

Here \(\mathcal{Q}\) may denote a single request, a small incoming batch, or a rolling traffic window; Appendix Table~\ref{tab:appendix-tfidf-window} evaluates the same fixed rule across these settings.

The staged extension keeps the same retriever but changes the retrieval unit. A query can follow the identity path or be decomposed into sub-queries. Each sub-query is retrieved independently; candidate scores are aggregated across sub-queries; and the top-\(k\) tools under the aggregated score are passed to the reranker. The reranker must then match the pool it actually sees, because merged staged pools and direct pools follow different candidate distributions. The hierarchical extension serves a different purpose. It predicts a small set of manual or automatically induced skills before tool retrieval, and we evaluate it as an efficiency mechanism around the direct dense route.

\section{Experimental Setup}

We evaluate on two benchmarks. \textbf{ToolRet} is the main retrieval-stress benchmark. The controlled \texttt{1100}-query normalized \texttt{web} subset serves as the main diagnostic slice; a broader \texttt{4996}-query aggregate over \texttt{19} source-query generators tests same-corpus source-style sensitivity; and a full merged ToolRet build over \texttt{44,453} tools and \texttt{7,726} queries tests whether the same pattern survives at larger scale across the \texttt{web}, \texttt{code}, and \texttt{customized} corpora. \textbf{StableToolBench}, an existing tool-learning benchmark derived from ToolBench, plays a different role in our study. As retrieval on this benchmark starts near saturation, we use it mainly for efficiency and hierarchy analyses.

The off-the-shelf comparisons include BM25, \texttt{bge-small-en-v1.5}, and \texttt{BGE-M3}. We choose these baselines to cover a lexical retriever, a lightweight English dense encoder, and a stronger general-purpose dense encoder. Candidate depth is fixed at \texttt{20} for the main dense runs. Reranker datasets use deterministic query-hash train/validation/test splits, and all reranking metrics are reported on held-out candidate-covered test queries from that protocol. Appendix~\ref{app:metric-details} reports the exact split counts and stronger reranker variants, and Appendix~\ref{app:retriever-ft-details} reports additional backbone comparisons and the mechanism-level ToolRerank comparison.

When candidate generation can fail outright, we additionally report a coverage-weighted top-1 proxy,
{
\[
G = C_m \cdot P@1_{\mid \mathrm{cov}},
\]
}
where \(C_m\) is the top-\(m\) candidate coverage for the current evaluation split, and \(P@1_{\mid \mathrm{cov}}\) is top-1 accuracy computed only over covered queries, i.e., queries whose retrieved top-\(m\) pool contains at least one gold tool. Thus, \(G\) gives a global top-1 proxy: covered queries are evaluated by their conditional top-1 accuracy, and queries with no gold tool in the retrieved top-\(m\) pool are counted as failures.

\paragraph{Skill-card rendering.}
To separate source-style collapse from raw API-schema formatting, we also rerender each tool as an executable skill card. The card normalizes the visible representation into fields such as capability, when-to-use conditions, inputs, and output or effect, while keeping the underlying tool identity, gold labels, splits, retrievers, and routing policy unchanged. This experiment asks whether the same retrieval-gate failure persists under a more skill-like capability representation.

\section{Results}

\subsection{Off-the-Shelf Diagnosis}

Table~\ref{tab:main-offtheshelf} gives the main off-the-shelf two-stage reference: a direct dense retriever followed by the lightweight learned reranker. The table shows a clear difference between the two benchmarks. ToolRet has substantially lower candidate coverage, even on the controlled \texttt{1100}-query web slice, while StableToolBench starts near saturation. We therefore use ToolRet for the main source-style collapse analysis and StableToolBench primarily as a high-coverage efficiency setting. Stronger reranker variants and candidate-covered split details are reported in Appendix~\ref{app:metric-details}. 

\begin{table}[t]
\centering
\footnotesize
\setlength{\tabcolsep}{4pt}
\resizebox{\columnwidth}{!}{
\begin{tabular}{llccc}
\toprule
Benchmark & Setting & Cov. & \multicolumn{2}{c}{P@1} \\
\cmidrule(lr){4-5}
 &  &  & Sem. & Lrn. \\
\midrule
\multirow{2}{*}{ToolRet} & \texttt{bge-small} / full 1100 & 67.7\% & 25.0\% & 50.0\% \\
 & \texttt{BGE-M3} / full 1100 & \textbf{80.8\%} & \textbf{34.9\%} & \textbf{55.8\%} \\
\multirow{2}{*}{StableToolBench} & \texttt{bge-small} / full 765 & 97.3\% & \textbf{70.1\%} & \textbf{71.0\%} \\
 & \texttt{BGE-M3} / full 765 & \textbf{98.4\%} & 66.7\% & 68.5\% \\
\bottomrule
\end{tabular}
}
\caption{Off-the-shelf two-stage results under a shared direct retrieval pipeline. ``Sem.'' denotes retriever-only top-\texttt{1} accuracy, and ``Lrn.'' denotes top-\texttt{1} accuracy after the lightweight learned reranker. Coverage is top-\texttt{20} candidate coverage. The results show that ToolRet leaves substantially more candidate-generation headroom than StableToolBench under the same protocol. Full covered-query metrics are provided in Appendix~\ref{app:metric-details}.}
\label{tab:main-offtheshelf}
\end{table}


\subsection{Adaptation Sensitivity}

\begin{figure}[t]
\centering
\includegraphics[width=0.5\textwidth]{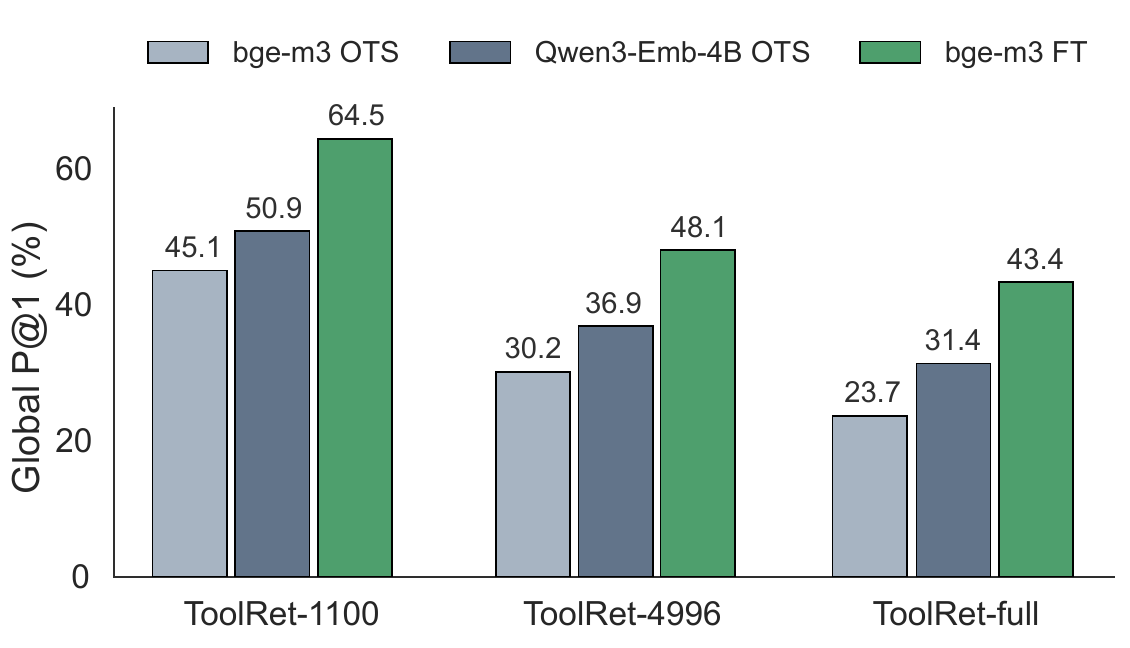}
\caption{Direct-dense retrieval adaptation across ToolRet-\texttt{1100}, ToolRet-\texttt{4996}, and ToolRet-full. The plotted score is the coverage-weighted top-1 proxy. A substantially larger 2025 off-the-shelf backbone (\texttt{Qwen3-Emb-4B}) narrows the gap in every ToolRet setting while leaving sizable room for source-adapted retrieval. Exact values are listed in Appendix~\ref{app:retriever-ft-details}.}
\label{fig:retriever-ft-summary}
\end{figure}


\paragraph{Matched-source adaptation improves the global top-1 proxy.}
We next evaluate whether adapting the retriever on matched query-tool supervision improves the coverage-weighted global top-1 proxy \(G\), which counts uncovered queries as failed top-\texttt{1} cases. Across the controlled ToolRet-\texttt{1100} slice, the broader ToolRet-\texttt{4996} aggregate, and the full ToolRet build, source-adapted \texttt{BGE-M3} consistently improves \(G\) by about \texttt{18--20} percentage points over the off-the-shelf retriever. A larger 2025 off-the-shelf backbone, Qwen3-Emb-4B~\citep{zhang2025qwen3embedding}, also improves \(G\), but it remains below the smaller source-adapted \texttt{BGE-M3} model across all three ToolRet settings. This means that model scale helps, while matched-source retriever adaptation still contributes a substantial additional gain under the same tool corpus. The same conclusion holds at \texttt{44,453} tools and \texttt{7,726} queries: full-corpus fine-tuning raises top-\texttt{20} candidate coverage by \texttt{23.5} percentage points with only a modest increase in average candidate tokens. Appendix~\ref{app:retriever-ft-details} reports the full numeric breakdown, bootstrap results, and additional backbone comparisons. A multimodal-capable LCO-Omni run in Table~\ref{tab:appendix-lco-omni} shows the same qualitative pattern with a frozen encoder and learned projection adapter.

StableToolBench serves as a boundary case. Nevertheless, matched-source fine-tuning is still positive: coverage rises from \texttt{96.3\%} to \texttt{99.1\%}, the learned dense top-1 score rises from \texttt{68.5\%} to \texttt{75.0\%}, and average candidate tokens decrease. The gain comes mainly from a cleaner candidate pool instead of a large uncovered-tail rescue, which is why we keep StableToolBench primarily for hierarchy and efficiency analysis (Appendix~\ref{app:retriever-ft-details}).

\paragraph{Transfer can catastrophically fail.}
A checkpoint trained only on the \texttt{1100}-query \texttt{toolbench}-style slice transfers poorly to the broader \texttt{4996}-query aggregate, where depth-\texttt{20} coverage collapses to \texttt{22.3\%}; by contrast, an aggregate-trained checkpoint still returns \texttt{87.3\%} coverage on the held-out \texttt{1100} test split. Appendix~\ref{app:retriever-ft-details} shows the most striking individual cases: the \texttt{1100}-trained checkpoint remains strong on the embedded \texttt{toolbench} source (\texttt{91.8\%} coverage) but collapses on \texttt{APIGen} (\texttt{0.7\%}), \texttt{ToolACE} (\texttt{0.0\%}), and \texttt{UltraTool} (\texttt{8.3\%}).

\paragraph{Lexical overlap fails to explain the failure.}
Panel~A of Figure~\ref{fig:toolscout-main} makes the paradox visible. Lexical cues alone fail to explain the collapse, because \texttt{APIGen} actually has \emph{higher} average query-to-gold-tool lexical overlap than the \texttt{toolbench} slice. A lexical retrieval control in Appendix Table~\ref{tab:appendix-lexical-fusion} sharpens that point. BM25 is competitive on high-overlap sources such as \texttt{APIGen} (\texttt{75.0\%} coverage) and \texttt{ToolACE} (\texttt{67.4\%}), and BM25+dense fusion usually improves further, but the gains vary by source, with dense still much stronger on \texttt{UltraTool} and the hybrid barely moving \texttt{toollens}. More importantly, in the actual collapsed \texttt{FT-1100} setting, BM25+\texttt{FT-1100} fusion raises the five-source aggregate from \texttt{1.6\%} to \texttt{54.2\%}, yet still trails BM25 alone at \texttt{59.4\%} across the same \texttt{3014} queries (Appendix Table~\ref{tab:appendix-lexical-fusion-ft1100}). This supports routing away from an unsafe dense specialist instead of blending it with a lexical branch.

\paragraph{TF-IDF flags clearly unsafe source styles.}
Across all \texttt{19} source-query configurations, TF-IDF centroid distance from the training source tracks transfer collapse more strongly than semantic-centroid, length, or multi-step proxies (query-count-weighted \texttt{r=-0.85}). As a leave-one-config-out compatibility routing criterion for clearly low-coverage transfer sources (\texttt{FT-1100} coverage below \texttt{30\%}), the same TF-IDF probe reaches \texttt{78.9\%} accuracy and \texttt{0.88} F1, ahead of semantic-centroid distance (\texttt{73.7\%}, \texttt{0.85}), query-length gap (\texttt{68.4\%}, \texttt{0.80}), and multi-step-rate gap (\texttt{47.4\%}, \texttt{0.58}).

A matched tool-side ablation over the normalized tool texts actually encoded by the retriever shows a related but weaker source fingerprint: tool-side TF-IDF distance still tracks the same collapse at weighted \texttt{r=-0.77}, and remains tightly coupled to the query-side distance (weighted \texttt{r=0.96}; Appendix~\ref{app:retriever-ft-details}). Appendix Table~\ref{tab:appendix-synthetic-rewrite} and \ref{tab:appendix-paired-synthetic} then sharpen the boundary: query-only rewrites shift the fingerprint without recreating the natural collapse, while paired synthetic source styles reproduce substantial negative transfer. Together these controls point to a broader source-style shift whose footprint remains visible on both sides of the retrieval pair, even though the query-side view is the stronger operational detector.

\subsection{Source-Aware Response Policy}

Panel~C of Figure~\ref{fig:toolscout-main} visualizes the first two stages of the response policy. On the mixed \texttt{4996}-query stream, always using the mismatched \texttt{FT-1100} checkpoint leaves coverage at only \texttt{22.3\%}. Applying the conservative TF-IDF rule \(\,d_{\text{tfidf}} < \tau\,\) to route high-coverage source styles to \texttt{FT-1100} and low-coverage transfer sources to the broader aggregate checkpoint raises coverage to \texttt{86.1\%}, essentially matching the fixed aggregate-trained retriever's \texttt{85.2\%}. In practice, the routing criterion preserves the narrow checkpoint on compatible traffic while routing away from severe coverage-loss cases, before aggregate-level supervision is necessarily available. Appendix Table~\ref{tab:appendix-mixed-ft} shows that naive mixed-training controls fail as a replacement: balanced and temperature-reweighted aggregate fine-tuning leave the collapsed sources unimproved while materially degrading compatible \texttt{toolbench}-style traffic. Two practical alternatives in Appendix Tables~\ref{tab:appendix-category-filter} and ~\ref{tab:appendix-lexical-fusion-ft1100} recover part of the lost coverage: oracle category filtering raises the mismatched checkpoint to \texttt{32.1\%}, and BM25+\texttt{FT-1100} fusion reaches \texttt{54.2\%} on the full five-source set. The routing criterion is useful when a system wants to retain a specialist on matching traffic, where \texttt{FT-1100} reaches \texttt{91.8\%} coverage compared with \texttt{87.3\%} for aggregate training.

The few-shot repair table reports the five natural ToolRet source slices where the \texttt{FT-1100} checkpoint loses coverage most severely in the mixed-source evaluation, including \texttt{reversechain}; these rows test whether a small amount of matched supervision can recover each collapsed source.

\begin{table}[t]
\centering
\footnotesize
\setlength{\tabcolsep}{4pt}
\resizebox{\columnwidth}{!}{
\begin{tabular}{lrrrrrr}
\toprule
& \multicolumn{2}{c}{\texttt{FT-1100}} & \multicolumn{2}{c}{\texttt{20}-shot} & \multicolumn{2}{c}{\texttt{FT-aggregate}} \\
\cmidrule(lr){2-3}\cmidrule(lr){4-5}\cmidrule(lr){6-7}
Source & Cov. & Global & Cov. & Global & Cov. & Global \\
\midrule
\texttt{APIGen} & 0.7\% & 0.3\% & 93.2\% & 74.4\% & 96.6\% & 73.5\% \\
\texttt{ToolACE} & 0.0\% & 0.0\% & 77.1\% & 46.0\% & 81.7\% & 47.2\% \\
\texttt{UltraTool} & 8.3\% & 5.7\% & 84.7\% & 48.1\% & 98.6\% & 81.9\% \\
\texttt{toollens} & 5.4\% & 1.3\% & 58.9\% & 27.1\% & 89.3\% & 47.3\% \\
\texttt{reversechain} & 5.1\% & 2.3\% & 89.7\% & 61.5\% & 79.5\% & 32.3\% \\
\bottomrule
\end{tabular}}
\caption{Multi-source few-shot repair from a collapsed \texttt{FT-1100} checkpoint. ``Global'' is the coverage-weighted multilayer perceptron (MLP) top-1 proxy. The table reports the headline \texttt{20}-shot setting only; the full \texttt{5/10/20/50}-shot sweep is listed in Appendix~\ref{app:retriever-ft-details}. The \texttt{reversechain} row has only \texttt{39} held-out test queries, so we treat its score cautiously. Sample complexity varies by source: the appendix sweep shows that \texttt{50}-shot continuation closes the \texttt{UltraTool} gap and narrows \texttt{toollens} further.}
\label{tab:fewshot-repair-main}
\end{table}

Table~\ref{tab:fewshot-repair-main} shows the next stage in that same procedure: once a small amount of matched supervision becomes available, short continuation fine-tuning can repair most of the remaining source-specific loss. Across these five collapsed sources, \texttt{20} matched examples raise the weighted MLP global proxy from \texttt{1.3\%} under the unusable \texttt{FT-1100} checkpoint to \texttt{53.9\%}, close to the \texttt{59.0\%} aggregate-trained reference. The full shot-count sweep in Appendix~\ref{app:retriever-ft-details} shows that \texttt{5}-shot repair generally fails, while \texttt{50}-shot continuation approaches the aggregate ceiling on harder sources. The remaining gaps on \texttt{UltraTool} and especially \texttt{toollens} therefore point to higher source-specific sample complexity: \texttt{50} matched examples close \texttt{UltraTool} to its aggregate reference, while \texttt{toollens} remains the hardest case even at \texttt{50} shots.

\subsection{Skill-Card Rendering Check}

\begin{table*}[t]
\centering
\footnotesize
\setlength{\tabcolsep}{5pt}
\begin{tabular}{lrrrrrr}
\toprule
Rendering & \texttt{1100} OTS & \texttt{1100} FT & \texttt{FT-1100}$\rightarrow$\texttt{4996} & \texttt{4996} OTS & \texttt{4996} FT & Switch \\
\midrule
Tool schema & 80.8\% & 91.8\% & 22.3\% & 73.7\% & 85.2\% & 86.1\% \\
Skill card & 72.2\% & 89.2\% & 21.7\% & 64.5\% & 84.0\% & 85.2\% \\
\bottomrule
\end{tabular}
\caption{Skill-card rendering check. Coverage is depth-\texttt{20} candidate coverage. The skill-card rendering keeps the same tools, labels, splits, retrievers, and routing policy, while rewriting visible tool text into normalized executable skill cards. Collapse and the switch policy persist under the more skill-like representation. Full per-source values and the deterministic card template are in Appendix~\ref{app:skillcard-rendering}.}
\label{tab:skillcard-main}
\end{table*}

To test whether source-style collapse is an artifact of raw API-schema rendering, we rerendered each tool as an executable skill card while keeping the tool identities, gold labels, splits, retrievers, and routing policy unchanged. Table~\ref{tab:skillcard-main} shows that the same pattern reappears. On the \texttt{1100} slice, skill-card rendering still benefits from matched-source fine-tuning, raising coverage from \texttt{72.2\%} to \texttt{89.2\%}. The same \texttt{FT-1100} checkpoint then collapses on the mixed \texttt{4996}-query stream, reaching only \texttt{21.7\%} coverage, while the aggregate-trained checkpoint reaches \texttt{84.0\%}. The collapsed sources remain severely affected, with coverage staying near zero for four of the five sources and reaching only \texttt{15.4\%} on \texttt{reversechain}; see Appendix Table~\ref{tab:appendix-skillcard-collapsed}. The detector also transfers: weighted TF-IDF correlation remains \texttt{-0.851}, ahead of semantic distance at \texttt{-0.576}, and the TF-IDF switch raises mixed coverage from \texttt{21.7\%} to \texttt{85.2\%}. This rules out raw schema formatting as the sole cause; the failure persists under a more skill-like executable capability representation.

\subsection{Depth and Extensions}

\begin{figure}[t]
\centering
\includegraphics[width=\linewidth]{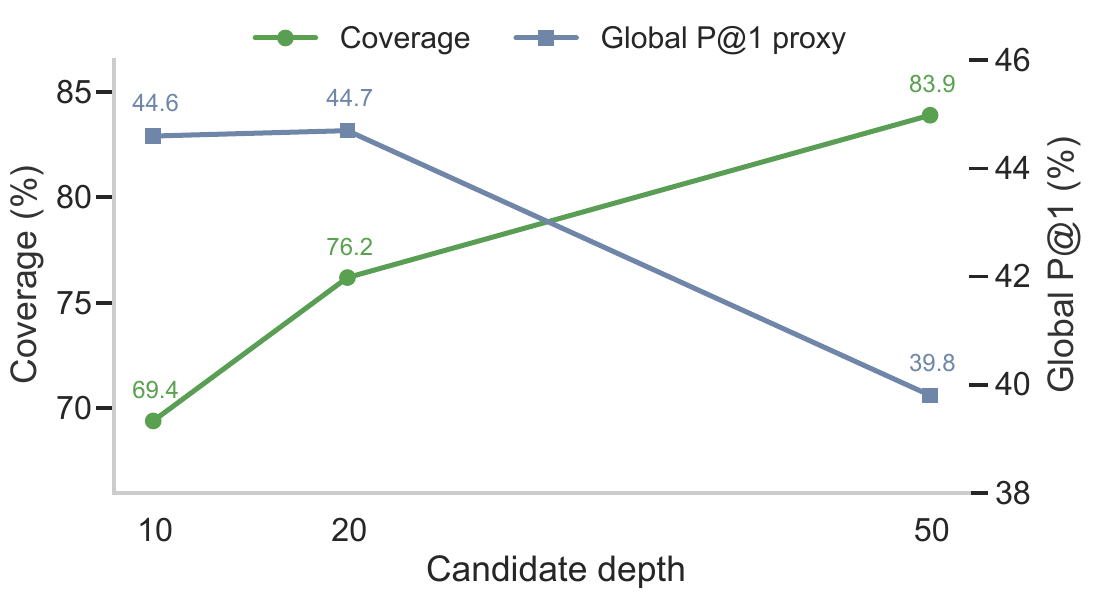}
\caption{ToolRet depth sweep on the full \texttt{1100}-query normalized web subset with \texttt{BGE-M3}. Coverage rises with depth, but the coverage-weighted top-1 proxy peaks around the main \texttt{20}-candidate operating point and then declines. Axes use truncated ranges for readability. Exact values are listed in Appendix~\ref{app:depth-details}.}
\label{fig:depth-tradeoff}
\end{figure}

The remaining analyses support the main collapse result as targeted follow-ups. Figure~\ref{fig:depth-tradeoff} shows that increasing candidate depth has a cost: raising depth from \texttt{10} to \texttt{50} lifts coverage from \texttt{69.4\%} to \texttt{83.9\%}, but the global proxy peaks at depth \texttt{20} and then declines. The remaining failures are also structured: under the strongest direct dense pipeline, \texttt{152/210} uncovered queries fall into the \texttt{many\_tools} or \texttt{broad\_web\_aggregation} buckets, which identifies the compositional multi-intent tail as the main residual failure mode.

That tail motivates staged retrieval as a conditional extension. On the main \texttt{1100}-query slice, split-and-merge retrieval raises overall coverage from \texttt{76.2\%} to \texttt{82.9\%} and improves the two main compositional buckets by about six points. Once the reranker is retrained on the merged candidate distribution, the staged global proxy rises to \texttt{49.8\%} with an adapted MLP and to \texttt{55.6\%} with a compact cross-encoder. Direct matched-source retriever fine-tuning still remains the stronger default move (\texttt{64.5\%}), so we use staged routing as a targeted repair for the compositional tail. A compact follow-up table is reported in Appendix~\ref{app:decomposition-details}.

Hierarchy serves as a coarse capability-layer extension. On StableToolBench, automatic taxonomies at the balanced \texttt{k=5} operating point preserve \texttt{89.2\%} coverage while reducing candidate context by \texttt{87.4\%}; Appendix~\ref{app:taxonomy-details} shows the corresponding token stakes, with benchmark-native full-catalog exposure at about \texttt{172k} tokens versus about \texttt{407} for retrieved top-\texttt{5} exposure. The paired benchmark-native execution check on \texttt{259} runnable queries across all six StableToolBench groups removes the quality-efficiency trade-off: under a fixed local \texttt{Qwen-14B} planner and a function-call-accuracy (FAC)-style judge, the official benchmark-provided exposure baseline solves \texttt{31.7\%} (\texttt{82/259}) while the automatic skill-tool exposure policy (\texttt{auto\_skill\_tool})  reaches \texttt{32.8\%} (\texttt{85/259}). This supports the broader capability-retrieval view while keeping the main reliability claim at the exact tool-backed skill retrieval gate.

\section{Conclusion}

This paper identifies source-style collapse as a same-corpus failure mode in tool retrieval: a retriever adapted to one source-defined slice can lose coverage on another slice over the same tool corpus. ToolScout addresses this failure with a query-side TF-IDF compatibility predictor, which routes likely mismatched traffic to an aggregate-trained checkpoint and then applies short continuation fine-tuning when matched examples become available. The skill-card experiment shows that the failure is not explained solely by raw API-schema formatting. Overall, the results show that reliable tool-using agents need reliable capability retrieval before reranking, planning, or execution can be effective.


\section*{Limitations}

Our results should be interpreted under four main limitations.

First, several comparisons use a shared retrieval, exposure, and downstream-selection pipeline. This design makes retrieve-then-rerank, tool-only routing, and hierarchy variants comparable at the routing layer, but it does not reproduce each system's original prompt stack, execution policy, or model choices. The results therefore support claims about the routing mechanism under a shared protocol, rather than full reproductions of the original agent stacks.

Second, the most detailed ToolRet reranking, depth, and failure-bucket analyses focus on the controlled \texttt{1100}-query normalized web subset. We repeat the core retriever-adaptation comparison on the full \texttt{7.6k}-task ToolRet build and observe the same large positive trend, but the merged \texttt{web}, \texttt{code}, and \texttt{customized} corpora provide less detailed error annotation. A positive replication on another independently generated multi-source tool corpus remains an important next test.

Third, the skill-card rendering result is limited to structured tools and APIs rendered as executable skill cards. Broader capability corpora with procedural skills, non-tool resources, and end-task skill incorporation remain outside the current study. The few-shot repair result is also scoped by supervision availability: matched examples could come from failed-query telemetry or tool-provider onboarding, but the rate at which such supervision accumulates in real deployments is left for future work.

Fourth, hierarchical routing is evaluated most strongly on StableToolBench. Our first ToolRet extension shows that manual categories remain stronger than automatic clusters, although both hierarchy variants substantially reduce candidate context. This suggests that category metadata quality matters, and broader tests on category-structured tool corpora are needed.

\bibliography{references}

\clearpage
\appendix
\section{Appendix}

This appendix is organized as follows. Appendix~\ref{app:metric-details}--\ref{app:retriever-ft-details} covers reranker checks, larger-scope ToolRet runs, and retriever adaptation details. Tables~\ref{tab:appendix-retriever-sources}--\ref{tab:appendix-mcp-invariance} collect source diagnostics, synthetic controls, MetaTool, skill-card rendering, and MCP-lite normalization. Appendix~\ref{app:depth-details}--\ref{app:taxonomy-details} covers depth, staged retrieval, hierarchy, and local execution follow-ups. Readers focused only on the main collapse claim can prioritize Tables~\ref{tab:appendix-reranker-stronger}, \ref{tab:appendix-lco-omni}, \ref{tab:appendix-source-compatibility}, \ref{tab:appendix-tfidf-window}, \ref{tab:appendix-category-filter}, \ref{tab:appendix-joint-source}, \ref{tab:appendix-synthetic-rewrite}, \ref{tab:appendix-paired-synthetic}, \ref{tab:appendix-skillcard-core}, \ref{tab:appendix-mixed-ft}, and \ref{tab:appendix-representation-geometry}.

\subsection{Reranker Splits and Checks}
\label{app:metric-details}

Table~\ref{tab:main-offtheshelf} in the main paper gives a compact off-the-shelf dense-routing reference. Candidate coverage is computed on the full benchmark slice used for each row, while the fuller P@1, R@5, and nDCG@5 metrics are reported here on the held-out candidate-covered test split from the reranker protocol described in the experimental setup. For the main ToolRet \texttt{BGE-M3} row, this means the reranking metrics come from the \texttt{129} held-out candidate-covered test queries inside the full \texttt{1100}-query run.

The exact query-hash split counts for the main dense reranker datasets are:
\texttt{635 / 126 / 129} train / val / test queries for ToolRet (\texttt{210} uncovered queries skipped) and
\texttt{551 / 94 / 108} for StableToolBench (\texttt{12} uncovered queries skipped).

The learned reranker is linear, so its feature usage is easy to summarize. Table~\ref{tab:appendix-feature-importance} reports grouped absolute coefficient mass for the two main dense runs. The StableToolBench model is dominated by retrieval-side evidence, while the ToolRet model also leans heavily on metadata-structure features. In both cases, execution-feedback features carry zero weight in the main dense setting.

\begin{table}[h]
\centering
\footnotesize
\setlength{\tabcolsep}{3pt}
\begin{tabular}{lccccc}
\toprule
Benchmark & Ret. & Meta. & Lex. & Query & Exec. \\
\midrule
ToolRet & 48.2\% & 44.1\% & 7.7\% & 0.0\% & 0.0\% \\
StableToolBench & 71.2\% & 14.3\% & 12.7\% & 1.8\% & 0.0\% \\
\bottomrule
\end{tabular}
\caption{Grouped reranker coefficient mass. Ret. = retrieval scores, reciprocal-rank features, and candidate-presence indicators; Meta. = metadata-structure features; Lex. = lexical overlap; Query = query-length features; Exec. = execution-feedback features.}
\label{tab:appendix-feature-importance}
\end{table}

We also ran stronger-reranker controls to test whether the paper's diagnosis is an artifact of the linear model. Table~\ref{tab:appendix-reranker-stronger} compares the main linear reranker against a two-hidden-layer MLP trained on the same feature sets and data splits, and a compact cross-encoder trained directly on the same candidate-covered reranker datasets. Both stronger rerankers improve covered-query quality, especially on ToolRet. The central interpretation stays unchanged: these gains occur only after candidate generation has already succeeded, so uncovered-query failures remain binding.

\begin{table}[h]
\centering
\footnotesize
\setlength{\tabcolsep}{3pt}
\begin{tabular}{llccc}
\toprule
Benchmark & Reranker & P@1 & R@5 & nDCG@5 \\
\midrule
ToolRet & Linear & 55.8\% & 73.0\% & 63.2\% \\
ToolRet & MLP & \textbf{64.3\%} & \textbf{78.6\%} & \textbf{70.4\%} \\
ToolRet & Cross-enc. & 76.7\% & 93.1\% & 84.7\% \\
StableTB & Linear & 68.5\% & 78.8\% & 71.6\% \\
StableTB & MLP & \textbf{68.5\%} & \textbf{81.2\%} & \textbf{73.5\%} \\
StableTB & Cross-enc. & 85.2\% & 89.5\% & 85.8\% \\
\bottomrule
\end{tabular}
\caption{Stronger reranker controls on the main dense runs. Metrics are reported on the same held-out candidate-covered test splits as Table~\ref{tab:main-offtheshelf}. Cross-encoders substantially raise the covered-query ceiling, and they still operate only after candidate generation has already succeeded.}
\label{tab:appendix-reranker-stronger}
\end{table}

\subsection{Larger-Scope ToolRet Aggregate}
\label{app:aggregate-details}

To test whether the main ToolRet pattern survives beyond the \texttt{1100}-query normalized web slice, we also built a larger aggregate benchmark by merging \texttt{19} ToolRet query configurations over the same \texttt{web} tool corpus. The resulting aggregate contains \texttt{4996} deduplicated queries (\texttt{2242} single-tool and \texttt{2754} multi-step), and it is meaningfully harder upstream: under the same depth-\texttt{20} dense candidate-generation protocol, \texttt{1316} queries still fail to place any gold tool into the candidate pool, so effective candidate coverage falls to \texttt{73.7\%}. Table~\ref{tab:appendix-aggregate} shows that this larger scope lowers absolute reranking quality, but preserves the same qualitative ordering as the main ToolRet result: stronger rerankers help substantially once a query is covered, while the dominant difficulty remains upstream candidate generation under a broader and more heterogeneous query mixture.

\begin{table}[h]
\centering
\footnotesize
\setlength{\tabcolsep}{4pt}
\begin{tabular}{lccc}
\toprule
Reranker & P@1 & R@5 & nDCG@5 \\
\midrule
BM25 & 27.0\% & 43.0\% & 35.1\% \\
Semantic & 33.2\% & 57.4\% & 45.3\% \\
Linear & 35.2\% & 58.1\% & 47.0\% \\
MLP & 41.0\% & 69.5\% & 55.6\% \\
Cross-encoder & \textbf{70.0\%} & \textbf{89.6\%} & \textbf{79.9\%} \\
\bottomrule
\end{tabular}
\caption{ToolRet aggregate stress test on \texttt{4996} queries from \texttt{19} query configs. Metrics are reported on the held-out candidate-covered test split from the same reranker protocol as Table~\ref{tab:main-offtheshelf}. The larger aggregate is substantially harder upstream (\texttt{73.7\%} candidate coverage), and once candidates are present the same reranker ordering reappears.}
\label{tab:appendix-aggregate}
\end{table}

\subsection{Task-Adapted Dense Retriever Fine-Tuning}
\label{app:retriever-ft-details}

We also tested a direct repair to the main bottleneck by fine-tuning the dense retriever itself. Starting from \texttt{BGE-M3}, we train query-tool pairs with \textbf{MultipleNegativesRankingLoss} and retrieval-confusing hard negatives. This intervention is intentionally aligned with the paper's diagnosis: if candidate generation is the first-order failure mode, then the cleanest method variant is to improve the retriever before changing the reranker or hierarchy.

\paragraph{Main-paper summary rows.}
Figure~\ref{fig:retriever-ft-summary} in the main paper visualizes the exact ToolRet source slices that carry the main retrieval-adaptation argument. Table~\ref{tab:appendix-retriever-ft-summary} lists the corresponding values explicitly.

\begin{table}[h]
\centering
\footnotesize
\setlength{\tabcolsep}{4pt}
\resizebox{\columnwidth}{!}{
\begin{tabular}{llrrr}
\toprule
Source & Retriever & Cov. & P@1 & Global \\
\midrule
\multirow{3}{*}{ToolRet-1100}
& \texttt{BGE-M3} & 80.8\% & 55.8\% & 45.1\% \\
& \texttt{Qwen3-Emb-4B} & 87.3\% & 58.3\% & 50.9\% \\
& \texttt{BGE-M3} FT on 1100 & \textbf{91.8\%} & \textbf{70.3\%} & \textbf{64.5\%} \\
\midrule
\multirow{3}{*}{ToolRet-4996}
& \texttt{BGE-M3} & 73.7\% & 41.0\% & 30.2\% \\
& \texttt{Qwen3-Emb-4B} & 78.9\% & 46.8\% & 36.9\% \\
& \texttt{BGE-M3} FT on 4996 & \textbf{85.2\%} & \textbf{56.5\%} & \textbf{48.1\%} \\
\midrule
\multirow{3}{*}{ToolRet-full}
& \texttt{BGE-M3} & 56.4\% & 42.0\% & 23.7\% \\
& \texttt{Qwen3-Emb-4B} & 72.6\% & 43.2\% & 31.4\% \\
& \texttt{BGE-M3} FT on full & \textbf{79.9\%} & \textbf{54.3\%} & \textbf{43.4\%} \\
\bottomrule
\end{tabular}}
\caption{Exact values behind the main-paper retrieval-adaptation summary figure. ``Global'' is the coverage-weighted top-1 proxy.}
\label{tab:appendix-retriever-ft-summary}
\end{table}

\paragraph{Hard-negative mining ablation.}
Because dense fine-tuning is standard in IR, we also asked whether the \emph{specific} hard-negative policy matters for tool retrieval. Table~\ref{tab:appendix-retriever-negatives} keeps the same \texttt{BGE-M3 + MNRL} training protocol on the \texttt{1100} split but swaps the negative source. Retrieval-confusing semantic negatives are strongest, BM25 negatives are close, random negatives are weaker, and same-category negatives are materially worse. This is useful method evidence: \emph{retrieval-confusing} supervision appears to matter most for tool-query alignment.

\begin{table}[h]
\centering
\footnotesize
\setlength{\tabcolsep}{4pt}
\resizebox{\columnwidth}{!}{
\begin{tabular}{lrrrrr}
\toprule
Negatives & FT Cov. & $\Delta$ Cov. & FT Multi & FT Recall \\
\midrule
Semantic hard & \textbf{88.5\%} & \textbf{+16.2} & 89.3\% & \textbf{71.7\%} \\
Semantic + random & \textbf{88.5\%} & \textbf{+16.2} & \textbf{90.0\%} & 68.2\% \\
BM25 hard & 87.8\% & +15.5 & 89.3\% & 69.2\% \\
Random & 84.5\% & +12.2 & 85.7\% & 68.1\% \\
Same-category & 78.4\% & +6.1 & 79.3\% & 58.8\% \\
\bottomrule
\end{tabular}}
\caption{Controlled hard-negative ablation for \texttt{BGE-M3 + MNRL} on the \texttt{1100} validation split. All rows start from the same off-the-shelf depth-\texttt{20} base coverage of \texttt{72.3\%}. Retrieval-confusing semantic negatives are strongest, while same-category negatives are much weaker despite being semantically plausible distractors.}
\label{tab:appendix-retriever-negatives}
\end{table}

\begin{table}[h]
\centering
\footnotesize
\setlength{\tabcolsep}{4pt}
\resizebox{\columnwidth}{!}{
\begin{tabular}{llrrr}
\toprule
Train Source & Eval Source & Cov. & Multi & Avg Tok. \\
\midrule
Off-the-shelf & ToolRet-1100 val & 72.3\% & 72.9\% & 1104.4 \\
FT on 1100 & ToolRet-1100 val & \textbf{88.5\%} & \textbf{90.0\%} & \textbf{910.9} \\
FT on 1100 & ToolRet-1100 test & \textbf{91.8\%} & \textbf{92.0\%} & \textbf{930.6} \\
\midrule
Off-the-shelf & ToolRet-4996 test & 65.5\% & 63.5\% & 1194.4 \\
FT on 4996 & ToolRet-4996 test & \textbf{85.2\%} & \textbf{89.5\%} & \textbf{836.2} \\
FT on 4996 & Transfer to 1100 test & 87.3\% & 88.0\% & 830.3 \\
\bottomrule
\end{tabular}}
\caption{Depth-\texttt{20} candidate-generation coverage for task-adapted dense retrievers. The \texttt{1100}-trained model is strongest on matched data, while the aggregate-trained model generalizes across the broader \texttt{4996}-query setting.}
\label{tab:appendix-retriever-ft-coverage}
\end{table}

\paragraph{Mixed-training controls.}
A natural alternative to TF-IDF-based routing is to reweight source slices during aggregate fine-tuning. Table~\ref{tab:appendix-mixed-ft} tests two such controls at candidate generation, where uncovered gold tools are unrecoverable by reranking. Simple reweighting leaves the collapsed sources unimproved relative to natural aggregate fine-tuning, and both variants materially degrade compatible \texttt{toolbench}-style traffic. The likely reason is that uniform balancing oversamples extremely low-support sources while undersampling high-support sources, weakening the aggregate checkpoint's natural in-distribution alignment.

\begin{table}[h]
\centering
\footnotesize
\setlength{\tabcolsep}{4pt}
\resizebox{\columnwidth}{!}{
\begin{tabular}{lllll}
\toprule
Training mix & 4996 & 1100 transfer & Collapsed-5 & \texttt{toolbench} \\
\midrule
Natural aggregate & \textbf{85.2\%} & \textbf{87.3\%} & \textbf{89.3\%} & \textbf{87.3\%} \\
Balanced & 81.8\% & 74.1\% & 88.5\% & 74.1\% \\
Temp.-reweighted & 81.8\% & 76.6\% & 88.3\% & 76.6\% \\
\bottomrule
\end{tabular}
}
\caption{Naive mixed-training controls for aggregate retriever fine-tuning. All rows use the same \texttt{BGE-M3 + MNRL} protocol and the same total number of one-per-query training examples; only the source sampling distribution changes. Coverage is reported at candidate depth \texttt{20}. ``Collapsed-5'' is the query-count-weighted mean over \texttt{APIGen}, \texttt{ToolACE}, \texttt{UltraTool}, \texttt{toollens}, and \texttt{reversechain}.}
\label{tab:appendix-mixed-ft}
\end{table}

\paragraph{Representation-geometry probe.}
Table~\ref{tab:appendix-representation-geometry} checks whether the mixed-training control also changes the learned similarity geometry. The key quantity is the mean margin between the best gold-tool score and the strongest non-gold candidate score at depth \texttt{20}. The collapsed-source failure of \texttt{FT-1100} is visible as a large negative margin, while balanced and temperature-reweighted aggregate training leave the collapsed-source margin largely unchanged and instead reduce the \texttt{toolbench} margin relative to natural aggregate fine-tuning. A simple source-ID linear probe also stays similarly high for the aggregate, balanced, and temperature-reweighted checkpoints, so naive reweighting mainly weakens high-support compatible geometry.

\begin{table*}[h]
\centering
\footnotesize
\setlength{\tabcolsep}{3.5pt}
\begin{tabular}{lrrrrr}
\toprule
Retriever & \texttt{toolbench} Cov. & \texttt{toolbench} Margin & Collapsed-5 Cov. & Collapsed-5 Margin & Probe F1 \\
\midrule
Off-the-shelf & 73.4\% & -0.033 & 71.9\% & -0.026 & 82.2\% \\
\texttt{FT-1100} & \textbf{91.8\%} & \textbf{+0.018} & 2.5\% & -0.371 & 76.6\% \\
\texttt{FT-aggregate} & 87.3\% & -0.016 & \textbf{89.3\%} & \textbf{+0.003} & \textbf{91.3\%} \\
Balanced aggregate & 74.1\% & -0.044 & 88.5\% & -0.009 & 90.9\% \\
Temp.-reweighted aggregate & 76.6\% & -0.039 & 88.3\% & -0.002 & 91.0\% \\
\bottomrule
\end{tabular}
\caption{Representation-geometry probe for the mixed-training control. Margins are the mean best-gold score minus strongest non-gold score at candidate depth \texttt{20}; higher is better. Probe F1 is a cross-validated source-style linear classifier over query embeddings for the six displayed sources.}
\label{tab:appendix-representation-geometry}
\end{table*}

\begin{table}[h]
\centering
\footnotesize
\setlength{\tabcolsep}{4pt}
\resizebox{\columnwidth}{!}{%
\begin{tabular}{lrrrrrr}
\toprule
Source & \texttt{FT-1100} & \texttt{5}-shot & \texttt{10}-shot & \texttt{20}-shot & \texttt{50}-shot & \texttt{FT-aggregate} \\
\midrule
\texttt{APIGen} & 0.7\% & 0.7\% & --- & 93.2\% & --- & 96.6\% \\
\texttt{ToolACE} & 0.0\% & 0.0\% & 68.0\% & 77.1\% & 86.3\% & 81.7\% \\
\texttt{UltraTool} & 8.3\% & 8.3\% & 81.9\% & 84.7\% & 98.6\% & 98.6\% \\
\texttt{toollens} & 5.4\% & 5.4\% & 28.6\% & 58.9\% & 87.5\% & 89.3\% \\
\texttt{reversechain} & 5.1\% & 5.1\% & 84.6\% & 89.7\% & 84.6\% & 79.5\% \\
\bottomrule
\end{tabular}
}
\caption{Coverage sweep for lightweight matched-source repair from a collapsed \texttt{FT-1100} checkpoint. The main paper reports the headline \texttt{20}-shot setting; this table gives the full \texttt{5/10/20/50}-shot progression. Dashes for \texttt{APIGen} mark shot counts skipped in the initial pilot.}
\label{tab:appendix-retriever-ft-fewshot}
\end{table}

\begin{table}[h]
\centering
\footnotesize
\setlength{\tabcolsep}{4pt}
\resizebox{\columnwidth}{!}{
\begin{tabular}{llrrr}
\toprule
Eval Source & Retriever & Lin. P@1 & MLP P@1 & MLP Global \\
\midrule
ToolRet-1100 & Off-the-shelf & 55.8\% & 64.3\% & 51.9\% \\
& FT on 1100 & \textbf{70.3\%} & \textbf{69.7\%} & \textbf{63.9\%} \\
& FT on 4996 & 54.8\% & 55.5\% & 48.5\% \\
\midrule
ToolRet-4996 & Off-the-shelf & 35.2\% & 41.0\% & 30.2\% \\
& FT on 1100 & 36.9\% & 48.7\% & 10.9\% \\
& FT on 4996 & \textbf{56.3\%} & \textbf{56.5\%} & \textbf{48.1\%} \\
\bottomrule
\end{tabular}
}
\caption{Direct dense reranker pipelines under task-adapted dense retrievers. ``MLP Global'' is the coverage-weighted top-1 proxy using the appropriate candidate-generation coverage for each source. The \texttt{1100}-trained checkpoint is strongest on the main ToolRet slice, while the aggregate-trained checkpoint remains stable on the broader \texttt{4996}-query aggregate.}
\label{tab:appendix-retriever-ft-direct}
\end{table}

\begin{table}[h]
\centering
\footnotesize
\setlength{\tabcolsep}{4pt}
\resizebox{\columnwidth}{!}{%
\begin{tabular}{lrrrrrr}
\toprule
Retriever & Cov. & Multi & Avg Tok. & Lin. Global & MLP P@1 & MLP Global \\
\midrule
Off-the-shelf & 56.4\% & 62.8\% & 1334.7 & 20.8\% & 42.0\% & 23.7\% \\
FT on full & \textbf{79.9\%} & \textbf{86.6\%} & 1454.5 & \textbf{42.5\%} & \textbf{54.3\%} & \textbf{43.4\%} \\
\bottomrule
\end{tabular}
}
\caption{Full ToolRet scale test on the merged \texttt{44,453}-tool / \texttt{7,726}-query build spanning \texttt{web}, \texttt{code}, and \texttt{customized} corpora. The same task-adapted dense-retrieval conclusion survives at full scale: candidate coverage rises by \texttt{+23.5} points and the direct dense MLP global proxy rises from \texttt{23.7\%} to \texttt{43.4\%}, while average candidate tokens increase by only about \texttt{9\%}.}
\label{tab:appendix-retriever-ft-full}
\end{table}

\begin{table*}[h]
\centering
\footnotesize
\setlength{\tabcolsep}{4pt}
\begin{tabular}{lrrrrr}
\toprule
Retriever & Cov. & Avg Tok. & Lin. P@1 & MLP P@1 & MLP Global \\
\midrule
Off-the-shelf & 96.3\% & 1508.0 & 68.5\% & 68.5\% & 66.0\% \\
FT on StableToolBench & \textbf{99.1\%} & \textbf{1304.5} & \textbf{75.0\%} & \textbf{75.0\%} & \textbf{74.3\%} \\
\bottomrule
\end{tabular}
\caption{StableToolBench direct dense reranker pipeline under the same \texttt{BGE-M3 + MNRL} retriever fine-tuning protocol. Because the benchmark already starts at very high depth-\texttt{20} coverage, the gain is smaller than on ToolRet in coverage terms (\texttt{96.3\%} to \texttt{99.1\%}), but the learned reranker still improves from \texttt{68.5\%} to \texttt{75.0\%} P@1 and from \texttt{66.0\%} to \texttt{74.3\%} on the coverage-weighted global proxy, while average candidate tokens decrease.}
\label{tab:appendix-retriever-ft-stb}
\end{table*}

The especially low \texttt{10.9\%} global score for ``FT on \texttt{1100}'' evaluated on ToolRet-\texttt{4996} is the combined effect of two cross-source failures. First, candidate generation collapses from the aggregate off-the-shelf baseline's \texttt{73.7\%} coverage to \texttt{22.3\%}. Second, the surviving covered queries are themselves harder and less aligned with the \texttt{1100}-trained reranker than the covered queries on the matched slice. The covered-query MLP P@1 therefore stays non-trivial (\texttt{48.7\%}), but once that ordering quality is multiplied by the much smaller covered set, the global proxy falls sharply.

\begin{table}[h]
\centering
\footnotesize
\setlength{\tabcolsep}{4pt}
\begin{tabular}{llrr}
\toprule
Eval Source & Retriever & CE P@1 & CE Global \\
\midrule
ToolRet-1100 & Off-the-shelf & \textbf{76.7\%} & 61.9\% \\
ToolRet-1100 & FT on 1100 & 73.1\% & \textbf{67.1\%} \\
\midrule
ToolRet-4996 & Off-the-shelf & \textbf{70.0\%} & 51.6\% \\
ToolRet-4996 & FT on 4996 & 65.0\% & \textbf{55.4\%} \\
\bottomrule
\end{tabular}
\caption{Compact cross-encoder view of the same retriever fine-tuning intervention. Candidate-covered cross-encoder P@1 is already high before retriever adaptation, so the fine-tuned checkpoints leave that covered-query margin similar; the global proxy still improves because matched-source fine-tuning raises candidate coverage substantially before reranking.}
\label{tab:appendix-retriever-ft-crossenc}
\end{table}

\paragraph{Bootstrap significance.}
We also ran paired bootstrap over the held-out test queries underlying the global top-1 proxy. Table~\ref{tab:appendix-retriever-ft-bootstrap} shows that the direct fine-tuning gains remain large and cleanly separated from zero on both ToolRet source slices: the \texttt{1100} linear comparison improves by \texttt{+19.0} points with a 95\% interval of \texttt{[10.8, 27.2]}, and the \texttt{4996} MLP comparison improves by \texttt{+19.3} points with a 95\% interval of \texttt{[15.4, 23.1]}. These numbers are close to the full-slice global proxies in the main text, with small differences because the bootstrap is computed on held-out test queries instead of the full benchmark slice.

\begin{table}[h]
\centering
\footnotesize
\setlength{\tabcolsep}{4pt}
\resizebox{\columnwidth}{!}{
\begin{tabular}{lrrrr}
\toprule
Comparison & Baseline & Candidate & $\Delta$ & 95\% CI \\
\midrule
\texttt{1100} direct FT (Linear) & 45.6\% & 64.6\% & +19.0 & [10.8, 27.2] \\
\texttt{4996} direct FT (MLP) & 30.6\% & 49.9\% & +19.3 & [15.4, 23.1] \\
\bottomrule
\end{tabular}}
\caption{Paired bootstrap significance for the main direct dense fine-tuning gains. Means are computed over the held-out test queries underlying the global top-1 proxy; both improvements have $p<0.001$.}
\label{tab:appendix-retriever-ft-bootstrap}
\end{table}

To test whether this effect is tied too closely to \texttt{BGE-M3}, we repeated the same \texttt{1100}-slice direct-dense evaluation with a smaller \texttt{bge-small-en-v1.5} backbone. The absolute ceiling is lower, but the direction is the same: MNRL fine-tuning still raises depth-\texttt{20} candidate coverage and substantially improves both linear and MLP direct dense reranking. This capacity control suggests the intervention is broader than one specific encoder family or capacity point.

\begin{table}[h]
\centering
\footnotesize
\setlength{\tabcolsep}{4pt}
\resizebox{\columnwidth}{!}{
\begin{tabular}{lrrrrr}
\toprule
Backbone & Setting & Cov. & Lin. P@1 & MLP P@1 & MLP Global \\
\midrule
\texttt{bge-small} & Off-the-shelf & 67.7\% & 50.0\% & 60.5\% & 38.1\% \\
& FT on 1100 & \textbf{86.1\%} & \textbf{68.3\%} & \textbf{69.7\%} & \textbf{57.6\%} \\
\texttt{BGE-M3} & Off-the-shelf & 80.8\% & 55.8\% & 64.3\% & 51.9\% \\
& FT on 1100 & \textbf{91.8\%} & \textbf{70.3\%} & \textbf{69.7\%} & \textbf{63.9\%} \\
\bottomrule
\end{tabular}
}
\caption{Cross-backbone validation of the direct retriever fine-tuning effect on the \texttt{1100} ToolRet slice. Both \texttt{bge-small} and \texttt{BGE-M3} improve substantially under the same MNRL fine-tuning protocol, even though the larger backbone remains stronger in absolute terms.}
\label{tab:appendix-retriever-ft-backbones}
\end{table}

\paragraph{Additional off-the-shelf backbone comparisons.}
We also checked whether the main ToolRet gap could be closed simply by swapping in different untuned embedding models. Table~\ref{tab:appendix-retriever-ft-backbone-swaps} compares the paper's off-the-shelf \texttt{BGE-M3} reference against a weaker 2023 swap (\texttt{e5-large-v2}) and a much larger 2025 backbone (\texttt{Qwen3-Emb-4B}). The result is consistent across the three ToolRet source slices where we have comparable runs: stronger off-the-shelf backbones help, while a clear gap to source-adapted \texttt{BGE-M3} remains. We use these rows as a backbone-replacement control, with a broader 2024/2025 retriever bake-off outside scope.

\begin{table*}[h]
\centering
\footnotesize
\setlength{\tabcolsep}{4pt}
\begin{tabular}{llrrrr}
\toprule
Source & Backbone & Cov. & Avg Tok. & MLP P@1 & MLP Global \\
\midrule
\multirow{3}{*}{ToolRet-1100}
& \texttt{BGE-M3} & \textbf{80.8\%} & --- & \textbf{64.3\%} & \textbf{51.9\%} \\
& \texttt{Qwen3-Emb-4B} & 87.3\% & \textbf{924.6} & 58.3\% & 50.9\% \\
& \texttt{e5-large-v2} & 73.4\% & 1141.6 & 56.5\% & 41.5\% \\
\midrule
\multirow{2}{*}{ToolRet-4996}
& \texttt{BGE-M3} & 73.7\% & --- & 41.0\% & 30.2\% \\
& \texttt{Qwen3-Emb-4B} & \textbf{78.9\%} & \textbf{905.7} & \textbf{46.8\%} & \textbf{36.9\%} \\
\midrule
\multirow{3}{*}{ToolRet-full}
& \texttt{BGE-M3} & \textbf{56.4\%} & \textbf{1334.7} & \textbf{42.0\%} & \textbf{23.7\%} \\
& \texttt{Qwen3-Emb-4B} & 72.6\% & \textbf{1097.3} & 43.2\% & 31.4\% \\
& \texttt{e5-large-v2} & 52.0\% & 1337.1 & 38.7\% & 20.2\% \\
\bottomrule
\end{tabular}
\caption{Additional off-the-shelf backbone comparisons on ToolRet under the same direct dense MLP reranking pipeline. The \texttt{BGE-M3} rows reuse the paper's off-the-shelf reference coverage together with the appendix MLP reranker results from Table~\ref{tab:appendix-retriever-ft-direct}; the \texttt{Qwen3-Emb-4B} and \texttt{e5-large-v2} rows come from corresponding held-out direct-dense runs on the same sources. The comparison tests whether changing the untuned encoder family closes the gap; a full retriever bake-off is outside the scope of this paper.}
\label{tab:appendix-retriever-ft-backbone-swaps}
\end{table*}

We also tested a multimodal-capable embedding backbone, \texttt{LCO-Embedding-Omni-3B-2605}~\citep{lco2026omni}. This model uses a different interface from the sentence-transformer training path in the main \texttt{BGE-M3} runs, so we used its official text-embedding route and trained a shared projection adapter on top of frozen query and tool embeddings. Table~\ref{tab:appendix-lco-omni} shows the resulting source-style sensitivity. The \texttt{1100}-adapted projection reaches high matched-source coverage (\texttt{86.7\%} raw, \texttt{88.0\%} skill-card) and then drops to about \texttt{58\%} on the mixed \texttt{4996} stream. Aggregate adaptation returns to \texttt{83.3\%} on the mixed stream and about \texttt{88\%} on the five collapsed sources. The TF-IDF signal keeps the same direction as the main \texttt{BGE-M3} result: weighted correlations with \texttt{1100}-adapted coverage are \texttt{-0.712} for raw tool text and \texttt{-0.682} for skill-card text, stronger than semantic distance in both cases.

\begin{table*}[h]
\centering
\footnotesize
\setlength{\tabcolsep}{4pt}
\begin{tabular}{llrrrr}
\toprule
Rendering & Adapter training & \texttt{1100} test & \texttt{4996} test & Collapsed-5 & TF-IDF switch \\
\midrule
Tool schema & \texttt{1100} & 86.7\% & 57.8\% & 55.1\% & 83.6\% \\
Tool schema & \texttt{4996} & 85.4\% & 83.3\% & 88.5\% & --- \\
Skill card & \texttt{1100} & 88.0\% & 58.7\% & 57.4\% & 84.0\% \\
Skill card & \texttt{4996} & 83.5\% & 83.3\% & 88.1\% & --- \\
\bottomrule
\end{tabular}
\caption{LCO-Omni projection-adapter validation. Values are depth-\texttt{20} candidate coverage using official last-token text embeddings from \texttt{LCO-Embedding-Omni-3B-2605}. The backbone is frozen and a shared linear projection is trained over query/tool embeddings. The pattern matches the main story: matched-source adaptation is strong, transfer to the mixed stream drops sharply, and the TF-IDF switch restores aggregate-level coverage.}
\label{tab:appendix-lco-omni}
\end{table*}

The source-style collapse result is also analyzable. The \texttt{1100} and \texttt{4996} test splits differ less in raw query length than in \emph{source-style mix}: the main \texttt{1100} slice is longer and more consistently multi-step (\texttt{43.3} tokens, \texttt{94.9\%} multi-step), whereas the aggregate test split is shorter on average (\texttt{36.2} tokens, \texttt{53.0\%} multi-step) but mixes \texttt{19} different source-query configurations. Table~\ref{tab:appendix-retriever-sources} shows that the \texttt{1100}-trained checkpoint remains strong on the embedded \texttt{toolbench} slice inside the aggregate, then collapses on several dissimilar source-query configurations. The aggregate-trained checkpoint is consistently positive across the main high-count configs. This supports the updated interpretation in the main text: the observed transfer failure is visible first on the query side and reflects a broader source-style shift across the retrieval pair.

\begin{table}[h]
\centering
\footnotesize
\setlength{\tabcolsep}{4pt}
\begin{tabular}{lrrrr}
\toprule
Config & Qs. & Base & FT-1100 & FT-4996 \\
\midrule
toolbench & 158 & 73.4\% & \textbf{91.8\%} & 87.3\% \\
APIGen & 146 & 82.9\% & 0.7\% & \textbf{96.6\%} \\
ToolACE & 175 & 79.4\% & 0.0\% & \textbf{81.7\%} \\
UltraTool & 72 & 69.4\% & 8.3\% & \textbf{98.6\%} \\
toollens & 56 & 21.4\% & 5.4\% & \textbf{89.3\%} \\
reversechain & 39 & 74.4\% & 5.1\% & \textbf{79.5\%} \\
\bottomrule
\end{tabular}
\caption{Per-config candidate coverage on the \texttt{4996} aggregate test split. The \texttt{1100}-trained checkpoint remains strong on the embedded \texttt{toolbench} source, while the aggregate-trained checkpoint improves coverage across heterogeneous source-query generators.}
\label{tab:appendix-retriever-sources}
\end{table}

\begin{table}[h]
\centering
\footnotesize
\setlength{\tabcolsep}{4pt}
\resizebox{\columnwidth}{!}{
\begin{tabular}{lrrrrr}
\toprule
Config & Qs. & Avg Tok. & Multi-step & Lex.\ Overlap & FT-1100 Cov. \\
\midrule
toolbench & 158 & 43.3 & 94.9\% & 0.195 & 91.8\% \\
APIGen & 146 & 18.0 & 24.0\% & 0.427 & 0.7\% \\
ToolACE & 175 & 58.3 & 18.3\% & 0.196 & 0.0\% \\
UltraTool & 72 & 37.3 & 75.0\% & 0.157 & 8.3\% \\
\bottomrule
\end{tabular}
}
\caption{Simple query-side feature probe for the held-out aggregate test rows of the main source-query generators discussed in the paper. ``Lex. Overlap'' is the average fraction of query content tokens that also appear in the corresponding gold-tool names, descriptions, or tags. A separate full-source query-only TF-IDF classifier over the \texttt{12} source styles with at least \texttt{50} queries reaches \texttt{87.5\%} accuracy and \texttt{81.0} macro-F1, showing that the source footprint is recoverable from query-side features. The routing experiments still use the fixed centroid rule.}
\label{tab:appendix-retriever-source-features}
\end{table}

\begin{table}[h]
\centering
\footnotesize
\setlength{\tabcolsep}{4pt}
\begin{tabular}{lrrrr}
\toprule
Source & Qs. & BM25 & Dense & BM25+Dense \\
\midrule
\texttt{toolbench} & 1099 & 51.0\% & 60.1\% & \textbf{64.9\%} \\
\texttt{APIGen} & 1000 & 75.0\% & 77.9\% & \textbf{82.1\%} \\
\texttt{ToolACE} & 1000 & 67.4\% & 72.7\% & \textbf{77.8\%} \\
\texttt{UltraTool} & 500 & 41.8\% & \textbf{67.4\%} & 62.4\% \\
\texttt{toollens} & 314 & 20.4\% & 21.3\% & \textbf{21.7\%} \\
\texttt{reversechain} & 200 & 46.5\% & 61.0\% & \textbf{63.5\%} \\
\midrule
\texttt{collapsed\_5} & 3014 & 59.4\% & 67.4\% & \textbf{69.9\%} \\
\bottomrule
\end{tabular}
\caption{Lexical BM25, off-the-shelf dense \texttt{BGE-M3}, and BM25+dense reciprocal-rank fusion on key ToolRet source-query configurations under the same top-\texttt{20} candidate budget. BM25 is competitive on high-overlap sources such as \texttt{APIGen}, and fusion usually improves further, but the gains remain source-dependent and leave hard cases. We use this table as a lexical control for the source-style transfer analysis.}
\label{tab:appendix-lexical-fusion}
\end{table}

\begin{table}[h]
\centering
\footnotesize
\setlength{\tabcolsep}{3pt}
\begin{tabular}{lrrrr}
\toprule
Source & Qs. & BM25 & FT-1100 & BM25+FT \\
\midrule
\texttt{toolbench} & 1099 & 51.0\% & \textbf{94.7\%} & 91.3\% \\
\texttt{APIGen} & 1000 & \textbf{75.0\%} & 0.7\% & 68.3\% \\
\texttt{ToolACE} & 1000 & \textbf{67.4\%} & 0.0\% & 63.8\% \\
\texttt{UltraTool} & 500 & \textbf{41.8\%} & 6.2\% & 36.2\% \\
\texttt{toollens} & 314 & \textbf{20.4\%} & 2.5\% & 15.0\% \\
\texttt{reversechain} & 200 & \textbf{46.5\%} & 1.5\% & 42.0\% \\
\midrule
\texttt{collapsed\_5} & 3014 & \textbf{59.4\%} & 1.6\% & 54.2\% \\
\bottomrule
\end{tabular}
\caption{Lexical BM25 versus the actually collapsed \texttt{FT-1100} dense checkpoint, plus reciprocal-rank fusion between the two, under the same top-\texttt{20} candidate budget. In the real mismatched setting, naive lexical+dense fusion underperforms BM25 alone across all five collapsed sources, even though the same \texttt{FT-1100} checkpoint remains excellent on the matched \texttt{toolbench} slice. This is why the paper routes away from an unsafe dense specialist instead of blending it with a lexical branch.}
\label{tab:appendix-lexical-fusion-ft1100}
\end{table}

\begin{table}[h]
\centering
\footnotesize
\setlength{\tabcolsep}{4pt}
\begin{tabular}{lrr}
\toprule
Retriever & Original & Oracle category \\
\midrule
\texttt{FT-1100} & 22.3\% & 32.1\% \\
\texttt{FT-4996} & 85.2\% & 88.4\% \\
\bottomrule
\end{tabular}
\caption{Oracle category-filtered candidate coverage on the mixed \texttt{4996}-query stream. The filter restricts retrieval to tools sharing a gold category, so it should be read as an upper-bound category restriction. It reduces part of the candidate-space difficulty while leaving checkpoint compatibility differences visible.}
\label{tab:appendix-category-filter}
\end{table}

We also reran the lightweight post-hoc source-distance probe across \emph{all \texttt{19}} aggregate source-query configurations, using TF-IDF centroid distance from the \texttt{toolbench\_1100} query distribution as a coarse measure of query-side shift. On the full set, Pearson correlation with \texttt{FT-1100} coverage is \texttt{-0.61}; the same comparison is weaker for \texttt{BGE-M3} sentence-embedding centroid distance (\texttt{-0.31}), query-length gap (\texttt{-0.22}), multi-step-rate gap (\texttt{-0.29}), and the config's own off-the-shelf base coverage (\texttt{0.44}). The gap between \texttt{-0.61} and the stronger high-count slices is explained by the aggregate long tail: several source-query configs contribute only \texttt{3--8} held-out queries, so their empirical coverage values are highly discrete and high-variance. We therefore also report two additional slices in the artifact: the query-count-weighted TF-IDF correlation remains strong at \texttt{-0.85}, while the same weighted correlation is only \texttt{-0.58} for sentence-embedding centroids; over the \texttt{9} configs with at least \texttt{10} held-out queries, TF-IDF again stays stronger (\texttt{-0.84} vs.\ \texttt{-0.67}). We report the full per-config probe in the supplemental artifact and treat it as descriptive. It supports the same qualitative conclusion as the main text: brittle transfer is better explained by source-style shift than by length, keyword overlap, or a simple semantic-centroid proxy alone.

\begin{figure}[h]
\centering
\includegraphics[width=\linewidth]{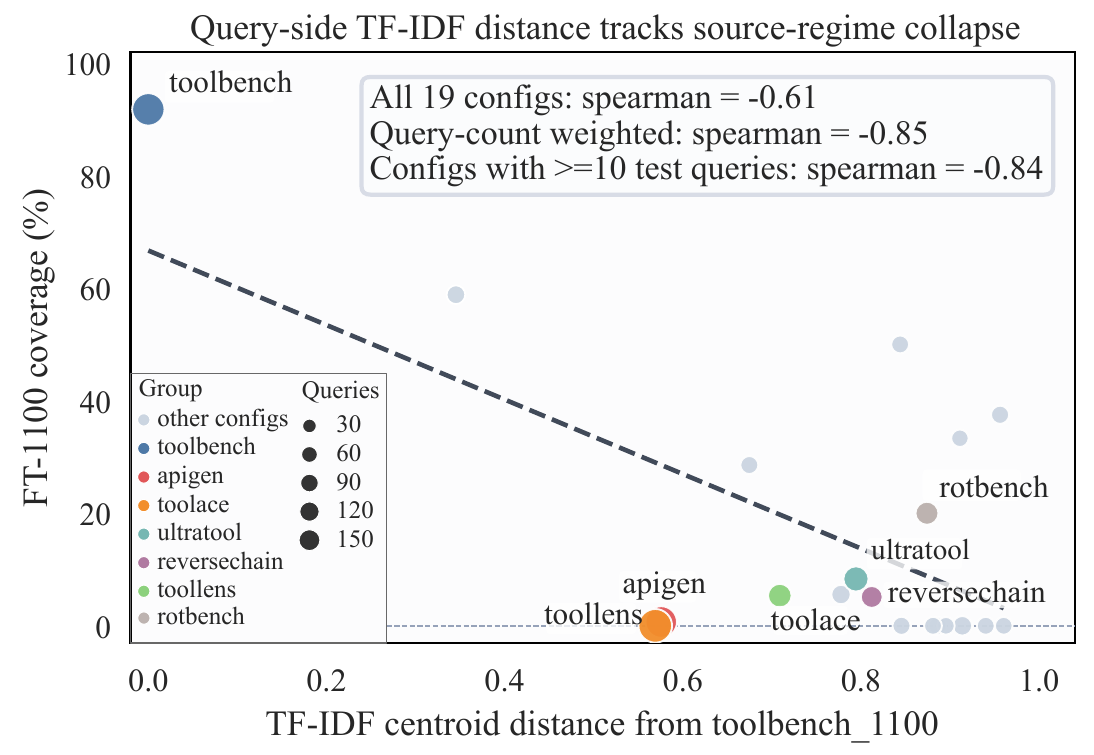}
\caption{Full \texttt{19}-config source-distance probe. Each point is one aggregate source-query configuration; point size reflects held-out query count. Across all configs the unweighted correlation is \texttt{r=-0.61}; weighting by query count gives \texttt{r=-0.85}; and restricting to configs with at least \texttt{10} held-out queries gives \texttt{r=-0.84}.}
\label{fig:source-distance}
\end{figure}

\begin{table}[h]
\centering
\footnotesize
\setlength{\tabcolsep}{4pt}
\resizebox{\columnwidth}{!}{
\begin{tabular}{lrrr}
\toprule
Proxy & LOO Acc. & 95\% Acc. CI & LOO F1 \\
\midrule
TF-IDF centroid distance & \textbf{78.9\%} & \texttt{[54.4, 93.9]} & \textbf{0.88} \\
\texttt{BGE-M3} centroid distance & 73.7\% & \texttt{[48.8, 90.9]} & 0.85 \\
Query-length gap & 68.4\% & \texttt{[43.4, 87.4]} & 0.80 \\
Multi-step-rate gap & 47.4\% & \texttt{[24.4, 71.1]} & 0.58 \\
\bottomrule
\end{tabular}
}
\caption{Leave-one-config-out compatibility check for unsafe transfer sources, defined as \texttt{FT-1100} coverage below \texttt{30\%}. F1 is reported for the positive \texttt{unsafe} class.}
\label{tab:appendix-source-compatibility}
\end{table}

Table~\ref{tab:appendix-source-compatibility} gives the exact leave-one-config-out compatibility check behind the main-paper summary sentence. Exact leave-one-config-out coverage regression is noticeably noisier: the weighted linear predictor built from TF-IDF centroid distance attains \texttt{0.17} mean absolute error, slightly better than the \texttt{BGE-M3} centroid-distance baseline (\texttt{0.18}), query-length gap (\texttt{0.21}), and multi-step-rate gap (\texttt{0.27}) but still too coarse for precise downstream coverage forecasting. The checker rests on only \texttt{19} source-level samples, so the exact accuracy intervals remain wide; we interpret source distance as a deployment-time \emph{compatibility routing criterion}.

\begin{table}[h]
\centering
\footnotesize
\setlength{\tabcolsep}{4pt}
\begin{tabular}{lrr}
\toprule
Window & Acc. & F1 \\
\midrule
\texttt{1--10} & 78.9\% & 0.882 \\
\texttt{20} & 83.1\% & 0.903 \\
\texttt{50--100} & 84.2\% & 0.909 \\
\bottomrule
\end{tabular}
\caption{Small-window application of the fixed TF-IDF compatibility rule. Each row uses sampled incoming windows and the same unsafe threshold as Table~\ref{tab:appendix-source-compatibility}.}
\label{tab:appendix-tfidf-window}
\end{table}

For latency, fitting the source centroid once takes \texttt{0.29} seconds on CPU. Across the recorded timing samples for transform-and-centroid computation on the same \texttt{4996}-query stream, the median wall-clock time is \texttt{388.7} ms and the empirical \texttt{95}th percentile is \texttt{452.5} ms. This timing covers the TF-IDF compatibility score itself, separate from dense retrieval.

To test whether this detector mainly sees the query side of the shift, we repeated the same analysis on the \emph{normalized tool texts actually encoded by the retriever}. Table~\ref{tab:appendix-joint-source} shows that tool-side TF-IDF distance remains informative but is weaker than the query-side probe: its weighted correlation with \texttt{FT-1100} coverage is \texttt{-0.77}, with \texttt{73.7\%} leave-one-config-out accuracy and \texttt{0.85} F1 for clearly low-coverage transfer sources. The query-side and tool-side distances are themselves tightly coupled (query-count-weighted \texttt{r=0.96}), and both are much larger cross-source than under matched-source split-half baselines (\texttt{0.15} for toolbench queries and \texttt{0.15} for the corresponding tool texts). Combining the two signals by averaging or taking a max leaves the detector unchanged, so the paper keeps the simpler query-side routing criterion in the main text. Mechanistically, this is broader \emph{source-style shift} across both sides of the retrieval pair.

\begin{table}[h]
\centering
\footnotesize
\setlength{\tabcolsep}{4pt}
\resizebox{\columnwidth}{!}{
\begin{tabular}{lrrr}
\toprule
Feature & Weighted \texttt{r} & LOO Acc. & LOO F1 \\
\midrule
Query-side TF-IDF & \textbf{-0.86} & \textbf{78.9\%} & \textbf{0.88} \\
Tool-side TF-IDF & -0.77 & 73.7\% & 0.85 \\
Average(query, tool) & -0.83 & \textbf{78.9\%} & \textbf{0.88} \\
Max(query, tool) & -0.86 & \textbf{78.9\%} & \textbf{0.88} \\
\bottomrule
\end{tabular}}
\caption{Query-side versus tool-side source fingerprints on ToolRet-\texttt{4996}. Tool-side distance is computed over the \emph{normalized tool texts actually encoded by the retriever} (name, description, arguments, examples) for each config's gold tools. The tool-side signal remains informative and tightly correlated with the query-side one, while the query-side detector stays strongest.}
\label{tab:appendix-joint-source}
\end{table}

\subsection{Additional Boundary Experiments}
\label{app:boundary-experiments}

We include three additional experiments to clarify when the ToolRet source-style effect does and does not appear. These experiments are not intended as new primary benchmarks. They test whether the failure can be reproduced by query-only rewriting, by paired synthetic source generation over the same tool corpus, or by task-family partitions in a second tool corpus.

\paragraph{Query-only rewrite control.}
We first test whether rewriting the query surface alone can reproduce the large source-style collapse observed on natural ToolRet sources. Using a local \texttt{Qwen2.5-14B-Instruct} rewriter, we converted the same \texttt{100} held-out \texttt{toolbench}-style queries into five target styles patterned after collapsed ToolRet sources. Table~\ref{tab:appendix-synthetic-rewrite} shows that the rewrite changes the query distribution: in TF-IDF space, it closes \texttt{37.8--142.0\%} of the natural target-source distance gap from \texttt{toolbench\_1100}, and it also moves in \texttt{BGE-M3} semantic space. The downstream coverage change is much smaller than the natural ToolRet collapse. The largest drop is the \texttt{ToolACE} rewrite, where \texttt{FT-1100} coverage changes from \texttt{92.0\%} to \texttt{83.0\%}; the other target styles leave \texttt{FT-1100} coverage between \texttt{89.0\%} and \texttt{93.0\%}. This suggests that query-only rewriting can partially move queries toward target-source styles, but the large natural collapse also depends on changes in the paired tool representations.

\begin{table}[h]
\centering
\footnotesize
\setlength{\tabcolsep}{3pt}
\resizebox{\columnwidth}{!}{
\begin{tabular}{lrrrr}
\toprule
Target & TF-IDF gap & Sem. gap & Off-the-shelf cov. & \texttt{FT-1100} cov. \\
\midrule
\texttt{APIGen} & 71.1\% & 30.7\% & 75.0 $\rightarrow$ 77.0 & 92.0 $\rightarrow$ 90.0 \\
\texttt{ToolACE} & 142.0\% & 87.2\% & 75.0 $\rightarrow$ 66.0 & 92.0 $\rightarrow$ 83.0 \\
\texttt{UltraTool} & 40.1\% & 10.3\% & 75.0 $\rightarrow$ 76.0 & 92.0 $\rightarrow$ 93.0 \\
\texttt{rotbench} & 52.1\% & 32.4\% & 75.0 $\rightarrow$ 72.0 & 92.0 $\rightarrow$ 89.0 \\
\texttt{reversechain} & 37.8\% & 23.0\% & 75.0 $\rightarrow$ 76.0 & 92.0 $\rightarrow$ 89.0 \\
\bottomrule
\end{tabular}
}
\caption{Synthetic query-only rewrite control on \texttt{100} held-out \texttt{toolbench}-style queries. Gap columns measure how much of the natural target-source centroid-distance gap from \texttt{toolbench\_1100} is closed by rewriting, under the same TF-IDF and \texttt{BGE-M3} semantic views used in the main source-style probe. Coverage columns report original-query $\rightarrow$ rewritten-query top-\texttt{20} coverage under off-the-shelf \texttt{BGE-M3} and the source-specific \texttt{FT-1100} retriever. Values above \texttt{100\%} indicate that the rewrite moves beyond the natural target source in that representation space.}
\label{tab:appendix-synthetic-rewrite}
\end{table}

\paragraph{Paired synthetic-source transfer.}
We next test whether transfer loss appears when both queries and tool renderings are generated under paired source styles over the same ToolRet tool corpus. Table~\ref{tab:appendix-paired-synthetic} reports two variants. The first uses the same local \texttt{Qwen2.5-14B-Instruct} generator with two prompts over the same candidate tool packs. The second keeps \texttt{Qwen2.5-14B-Instruct} for the natural task-description style and uses a separate \texttt{MirrorAPI} checkpoint for the compact API-template style. Both variants produce stronger transfer loss than query-only rewriting: fine-tuning improves matched-source coverage and loses coverage cross-source, with the cross-generator variant also falling below the off-the-shelf retriever in the mismatched direction.

\begin{table}[h]
\centering
\footnotesize
\setlength{\tabcolsep}{3.5pt}
\begin{tabular}{lrrrr}
\toprule
& \multicolumn{2}{c}{Qwen-only} & \multicolumn{2}{c}{Qwen/MirrorAPI} \\
\cmidrule(lr){2-3}\cmidrule(lr){4-5}
Train / Eval & Natural & API & Natural & API \\
\midrule
Off-the-shelf \texttt{BGE-M3} & 72.1\% & 56.3\% & 72.1\% & 39.1\% \\
FT natural & \textbf{89.5\%} & 60.6\% & \textbf{88.4\%} & 42.0\% \\
FT API-template & 55.8\% & \textbf{84.5\%} & 43.0\% & \textbf{82.6\%} \\
\bottomrule
\end{tabular}
\caption{Paired synthetic-source transfer over a shared ToolRet tool corpus. Values report top-\texttt{20} coverage. The Qwen-only variant uses the same \texttt{Qwen2.5-14B-Instruct} generator for both styles; the Qwen/MirrorAPI variant keeps the natural style fixed and uses \texttt{MirrorAPI} for the API-template style. Both settings show transfer loss under a shared tool corpus, with stronger loss in the cross-generator setting.}
\label{tab:appendix-paired-synthetic}
\end{table}

\paragraph{MetaTool task-family comparison.}
As a second-corpus comparison, we convert MetaTool into the same retrieval format and use its task-family labels as a coarse partition. This experiment asks whether arbitrary task-family differences are enough to reproduce ToolRet-style collapse. MetaTool's public task-family labels describe shared-process task families rather than independent query-generation sources, and the \texttt{subtask1\_similar} and \texttt{subtask3\_reliable} families share \texttt{100\%} of their queries. Table~\ref{tab:appendix-metatool-boundary} shows that family-specific fine-tuning does not produce ToolRet-style collapse: off-the-shelf \texttt{BGE-M3} already reaches \texttt{79.0\%} coverage overall, and all reported cross-family cells remain at \texttt{88.5\%} coverage or higher. This supports a bounded interpretation of the main result: the large ToolRet collapse is tied to independent source-generation effects over a shared tool corpus, rather than to any task partition within a tool benchmark.

\begin{table}[h]
\centering
\footnotesize
\setlength{\tabcolsep}{4pt}
\begin{tabular}{lrrr}
\toprule
Train $\downarrow$ / Eval $\rightarrow$ & Task1 & Similar & Scenario \\
\midrule
\texttt{task1} & 96.3\% & 91.0\% & 92.0\% \\
\texttt{similar} & 100.0\% & 91.7\% & 96.7\% \\
\texttt{scenario} & 93.8\% & 88.5\% & 100.0\% \\
\midrule
Off-the-shelf \texttt{BGE-M3} & 82.7\% & 78.8\% & 74.3\% \\
\bottomrule
\end{tabular}
\caption{MetaTool task-family comparison. Rows fine-tune \texttt{BGE-M3} on one single-tool task family and evaluate top-\texttt{20} coverage on another. MetaTool is useful as a second tool corpus, but its task-family labels are shared-process task labels. The absence of collapse here supports a bounded interpretation: ToolRet's source-style collapse is tied to independent generation sources more than arbitrary task-family differences inside a mostly shared generation process.}
\label{tab:appendix-metatool-boundary}
\end{table}

\clearpage

\subsection{Skill-Card Rendering Check}
\label{app:skillcard-rendering}

The main text reports a compact skill-card rendering check. This appendix gives the deterministic rendering template and the full coverage details. The check keeps tool identities, gold labels, train/test splits, retriever backbones, fine-tuning protocol, and routing policy fixed; only the visible tool text is rerendered. The template is:

\begin{table}[h]
\centering
\footnotesize
\setlength{\tabcolsep}{5pt}
\begin{tabular}{lp{0.68\columnwidth}}
\toprule
Field & Template content \\
\midrule
Skill & \texttt{\{tool\_name\}} \\
Capability & \texttt{This skill enables the agent to \{description\}.} \\
When to use & \texttt{Use this skill when the user request matches \{tags/categories/description\}.} \\
Inputs & \texttt{\{arguments or schema fields\}} \\
Output/effect & \texttt{\{available output or API-effect description\}} \\
Metadata & \texttt{\{tags and categories\}} \\
\bottomrule
\end{tabular}
\caption{Deterministic skill-card rendering template. The rendering is template-based and changes the visible capability representation while keeping labels, queries, and supervision fixed.}
\label{tab:appendix-skillcard-template}
\end{table}

Table~\ref{tab:appendix-skillcard-core} gives the main summary values. The original tool-schema row is copied from the main ToolRet experiments for comparison, while the skill-card row is retrained and evaluated on the rerendered corpora. Matched-source fine-tuning remains useful, the \texttt{FT-1100} transfer collapse remains severe, the aggregate checkpoint remains stable, and the same TF-IDF switch policy remains effective.

\begin{table}[h]
\centering
\footnotesize
\setlength{\tabcolsep}{3.5pt}
\resizebox{\columnwidth}{!}{
\begin{tabular}{lrrrrrr}
\toprule
Rendering & 1100 OTS & 1100 FT & FT-1100$\rightarrow$4996 & 4996 OTS & 4996 FT & Switch \\
\midrule
Tool schema & 80.8\% & 91.8\% & 22.3\% & 73.7\% & 85.2\% & 86.1\% \\
Skill card & 72.2\% & 89.2\% & 21.7\% & 64.5\% & 84.0\% & 85.2\% \\
\bottomrule
\end{tabular}}
\caption{Core skill-card rendering check. Values are depth-\texttt{20} candidate coverage. Rerendering tools as executable skill cards lowers the off-the-shelf baseline but preserves the main transfer-collapse and switch-policy pattern.}
\label{tab:appendix-skillcard-core}
\end{table}

Table~\ref{tab:appendix-skillcard-collapsed} shows the collapsed sources under skill-card rendering. The same sources remain severely affected by the \texttt{FT-1100} specialist, while aggregate training recovers high coverage.

\begin{table}[h]
\centering
\footnotesize
\setlength{\tabcolsep}{4pt}
\resizebox{\columnwidth}{!}{
\begin{tabular}{lrrrr}
\toprule
Source & Queries & OTS & \texttt{FT-1100} & \texttt{FT-aggregate} \\
\midrule
\texttt{APIGen} & 146 & 82.9\% & 2.1\% & 96.6\% \\
\texttt{ToolACE} & 175 & 75.4\% & 0.6\% & 85.7\% \\
\texttt{UltraTool} & 72 & 66.7\% & 1.4\% & 98.6\% \\
\texttt{toollens} & 56 & 23.2\% & 0.0\% & 87.5\% \\
\texttt{reversechain} & 39 & 66.7\% & 15.4\% & 82.0\% \\
\bottomrule
\end{tabular}}
\caption{Collapsed-source coverage under skill-card rendering. The natural collapsed sources remain collapsed under \texttt{FT-1100}, supporting the interpretation that raw schema formatting alone fails to explain the failure.}
\label{tab:appendix-skillcard-collapsed}
\end{table}

The detector result also persists. With skill-card rendering, weighted TF-IDF distance correlates with \texttt{FT-1100} coverage at \texttt{r=-0.851}, while weighted semantic distance reaches \texttt{r=-0.576}. A one-threshold TF-IDF switch raises mixed \texttt{4996} coverage from \texttt{21.7\%} under \texttt{FT-1100} to \texttt{85.2\%}; adding a base fallback gives \texttt{86.0\%}. This is a representation control for tool-backed executable skills; broader skill retrieval remains outside the scope of this experiment.

Finally, we tested a stricter schema-normalization counterfactual motivated by emerging tool standards such as MCP. Starting from the exact tool texts consumed by the retriever, we rebuilt both ToolRet-\texttt{1100} and ToolRet-\texttt{4996} under an \emph{MCP-lite} rendering that keeps only \texttt{name}, a cleaned \texttt{description}, and an \texttt{arguments}-style field with fixed ordering, while removing source-specific metadata and boilerplate. Table~\ref{tab:appendix-mcplite} shows that this paired retrain/eval ablation leaves the failure mode in place. The narrow-checkpoint transfer collapse remains severe (\texttt{22.3\%} $\rightarrow$ \texttt{25.1\%}), while off-the-shelf retrieval becomes noticeably weaker in absolute terms. We read this as evidence for a structural tradeoff: stricter standardization reduces some source-specific variation, yet it also discards surface cues that a default retriever exploits.

\begin{table}[h]
\centering
\footnotesize
\setlength{\tabcolsep}{4pt}
\begin{tabular}{lrrr}
\toprule
Setting & Original & MCP-lite & $\Delta$ \\
\midrule
\texttt{1100} off-the-shelf & 80.8\% & 68.4\% & -12.4 \\
\texttt{1100} FT (matched) & 91.8\% & 90.5\% & -1.3 \\
\texttt{FT-1100} $\rightarrow$ \texttt{4996} & 22.3\% & 25.1\% & +2.8 \\
\midrule
\texttt{4996} off-the-shelf & 73.7\% & 66.8\% & -6.9 \\
\texttt{4996} FT (matched) & 85.2\% & 84.3\% & -0.9 \\
\texttt{FT-4996} $\rightarrow$ \texttt{1100} & 87.3\% & 87.3\% & +0.0 \\
\bottomrule
\end{tabular}
\caption{Paired retrain/eval under a stricter MCP-lite tool rendering. Both ToolRet-\texttt{1100} and ToolRet-\texttt{4996} are rebuilt with unified \texttt{name + description + arguments} fields and the retrievers are retrained on the corresponding normalized corpora before evaluation. Schema normalization leaves the narrow-checkpoint collapse in place and lowers off-the-shelf coverage substantially. All values are retriever coverage, before reranked global top-1.}
\label{tab:appendix-mcplite}
\end{table}

A complementary representation-level check helps explain why the aggregate-trained checkpoint remains a stable fallback under the same normalization. Using the paired original/MCP-lite corpora above, we sampled \texttt{200} gold tools from each collapsed source (\texttt{1,000} tools total) and measured cosine similarity between the tool embeddings produced by the same retriever on the original versus MCP-lite text. Table~\ref{tab:appendix-mcp-invariance} shows a large and directionally consistent gap: the aggregate-trained checkpoint is much more stable than \texttt{FT-1100} across all five sources. The \texttt{APIGen} row is the most extreme because its source format is already close to the MCP-lite template, but the mean gap remains substantial even after excluding \texttt{APIGen}.

\begin{table}[h]
\centering
\footnotesize
\setlength{\tabcolsep}{4pt}
\begin{tabular}{lrrr}
\toprule
Source & \texttt{FT-1100} & \texttt{FT-aggregate} & $\Delta$ \\
\midrule
\texttt{APIGen} & 0.337 & 0.924 & +0.587 \\
\texttt{ToolACE} & 0.341 & 0.621 & +0.279 \\
\texttt{UltraTool} & 0.454 & 0.915 & +0.461 \\
\texttt{toollens} & 0.470 & 0.747 & +0.277 \\
\texttt{reversechain} & 0.451 & 0.705 & +0.254 \\
\midrule
Mean (all 5) & 0.411 & 0.782 & +0.372 \\
Mean (excl.\ \texttt{APIGen}) & 0.429 & 0.747 & +0.318 \\
\bottomrule
\end{tabular}
\caption{Representation-level invariance under MCP-lite normalization on collapsed-source gold tools. Each row reports the mean cosine similarity between the embedding of a tool's original text and the embedding of its MCP-lite text under the same retriever checkpoint, using \texttt{200} sampled gold tools per source (\texttt{1,000} tools total). The aggregate-trained checkpoint produces more stable tool representations than \texttt{FT-1100} in every source. The especially high \texttt{APIGen} value reflects that its source format is already close to the MCP-lite template.}
\label{tab:appendix-mcp-invariance}
\end{table}

\subsection{Depth Sweep Details}
\label{app:depth-details}

Figure~\ref{fig:depth-tradeoff} in the main paper compresses the candidate-depth tradeoff into coverage and global top-1. Table~\ref{tab:appendix-depth-sweep} gives the full covered-query and global metrics for the same sweep.

\begin{table}[h]
\centering
\footnotesize
\setlength{\tabcolsep}{6pt}
\resizebox{\columnwidth}{!}{
\begin{tabular}{ccccccc}
\toprule
Depth & Cov. & \multicolumn{2}{c}{P@1} & \multicolumn{2}{c}{R@5} & nDCG@5 \\
\cmidrule(lr){3-4}\cmidrule(lr){5-6}
 &  & Covered & Global & Covered & Global & Covered \\
\midrule
10 & 69.4\% & \textbf{64.3\%} & 44.6\% & \textbf{90.7\%} & \textbf{62.9\%} & \textbf{77.1\%} \\
20 & 76.2\% & 58.7\% & \textbf{44.7\%} & 78.7\% & 60.0\% & 67.6\% \\
50 & \textbf{83.9\%} & 47.4\% & 39.8\% & 64.8\% & 54.4\% & 55.2\% \\
\bottomrule
\end{tabular}
}
\caption{Exact ToolRet depth-sweep values behind Figure~\ref{fig:depth-tradeoff}. Increasing depth raises coverage, but covered-query quality decays enough that the global top-1 proxy peaks at the main \texttt{20}-candidate operating point.}
\label{tab:appendix-depth-sweep}
\end{table}

\subsection{Qualitative Case Studies}
\label{app:qualitative-cases}

Table~\ref{tab:appendix-cases} grounds the two main qualitative claims in concrete queries. The first row shows the same-corpus source-style shift behind the \texttt{APIGen} collapse: the gold tool names are almost literally present in the query, yet the phrasing follows a compact API-template style instead of the longer task-description supervision seen on the main \texttt{toolbench}-style slice. The second row shows the compositional tail that motivates staged retrieval: a single Yosemite request bundles weather lookup, trail discovery, and recommendation-style aggregation, and the conservative splitter rewrites it into two much cleaner retrieval problems before merged-pool reranking.

\begin{table*}[t]
\centering
\footnotesize
\setlength{\tabcolsep}{4pt}
\begin{tabular}{p{0.13\textwidth}p{0.21\textwidth}p{0.29\textwidth}p{0.27\textwidth}}
\toprule
Phenomenon & Query Form & Gold Tools & Why It Matters \\
\midrule
Same-corpus source-style shift &
\texttt{APIGen}: ``Please provide the latest songs from the artist Beyoncé and also list the countries in ascending order.'' &
\texttt{all\_songs\_from\_artist}; \texttt{getallcountry} &
The query has obvious lexical access to the tool schemas, yet it is phrased as a compact API-style command instead of a longer end-user task description. This is the kind of high-overlap cross-source case where the \texttt{1100}-trained retriever still collapses. \\
\midrule
Compositional tail / staged repair &
Yosemite query: weather forecast + trail recommendations + scenic-view filtering in one request &
\begin{tabular}[t]{@{}l@{}}
\texttt{VALUE\_SERP\_Google\_Search};\\
\texttt{VALUE\_SERP\_Google\_Place}\\
\texttt{\_Details};\\
\texttt{VALUE\_SERP\_Google\_Video}
\end{tabular} &
Single-shot retrieval must compress heterogeneous intents into one embedding. The conservative splitter rewrites this into ``Provide the weather forecast for Yosemite ...'' and ``Suggest popular hiking trails in Yosemite ...'', which produces two cleaner retrieval problems before the candidates are merged. \\
\bottomrule
\end{tabular}
\caption{Two concrete cases behind the main findings. The first illustrates why same-corpus source-style shift goes beyond missing keywords. The second illustrates why staged retrieval helps the compositional tail even when direct dense retrieval is already strong on average.}
\label{tab:appendix-cases}
\end{table*}

The staged pipeline reacts differently. Replacing the base retriever with the aggregate-trained checkpoint inside the conservative \texttt{Qwen-14B} staged pipeline improves the larger aggregate only marginally (\texttt{33.2\%} to \texttt{33.5\%} under the staged MLP) and degrades the held-out \texttt{1100} staged evaluation (\texttt{58.9\%} to \texttt{22.2\%}). These rows already use rerankers retrained on the corresponding staged merged-pool datasets, so the weak transfer is deeper than an unfitted second stage. This reinforces a separate point already visible in the main text: staged routing imposes its own merged-pool reranking problem, and a stronger retriever alone leaves the second-stage adaptation issue unsolved.

\subsection{Split-and-Merge Follow-Up Metrics}
\label{app:decomposition-details}

The main text reports the split-and-merge probe first as a coverage intervention and then as a staged-retrieval baseline with adapted rerankers. Here we report the same story in compact form. All rows below use the same \texttt{semantic + semantic\_rerank} retrieval setup and the same fixed \texttt{40}-candidate budget as the main split-and-merge probe; only the reranker changes.

\begin{table*}[h]
\centering
\footnotesize
\setlength{\tabcolsep}{4pt}
\begin{tabular}{lrrrr}
\toprule
Setting & Cov. & P@1 & R@5 & Global top-1 \\
\midrule
Single + original reranker & 76.2\% & 45.6\% & 77.5\% & 34.7\% \\
Split + original reranker & 82.9\% & 39.2\% & 77.3\% & 32.5\% \\
Split + adapted linear & 82.9\% & 46.8\% & 78.8\% & 38.8\% \\
Split + adapted MLP & \textbf{82.9\%} & \textbf{55.7\%} & \textbf{83.6\%} & \textbf{46.2\%} \\
Split + adapted cross-encoder & 82.9\% & 67.1\% & 93.1\% & 55.6\% \\
\bottomrule
\end{tabular}
\caption{Staged-retrieval follow-up on ToolRet under the same candidate-generation setup as the main split-and-merge probe. The merged pool hurts top-1 quality under the unchanged reranker, but adapted rerankers recover and then surpass the single-query global proxy. A compact cross-encoder pushes the same staged pool further upward, showing that the staged candidate generator remains competitive under stronger rerankers as well.}
\label{tab:appendix-decomposition}
\end{table*}

We keep the broader staged-routing family compact here. Learned decomposition variants and adaptive first-stage triggers were directionally consistent with the main split-and-merge result, and staged retrieval remains a conditional extension around the main collapse result.

\subsection{Hierarchy Efficiency and Benchmark-Native Execution}
\label{app:taxonomy-details}

Automatic taxonomies are induced with \textbf{MiniBatchKMeans}~\citep{pedregosa2011sklearn} over tool embeddings. In the main StableToolBench result, we use \texttt{46} clusters to match the benchmark's \texttt{46} manual categories, then summarize each cluster with its nearest representative tools and use those clusters as the first-stage skill space.

\begin{table*}[t]
\centering
\footnotesize
\setlength{\tabcolsep}{5pt}
\resizebox{.9\textwidth}{!}{%
\begin{tabular}{crrrrrrrr}
\toprule
\multirow{2}{*}{top-$k$ skills} & \multicolumn{4}{c}{Manual skill $\rightarrow$ tool} & \multicolumn{4}{c}{Automatic skill $\rightarrow$ tool} \\
\cmidrule(lr){2-5}\cmidrule(lr){6-9}
 & P@1 & R@5 & Cov. & Cand.\ Red. & P@1 & R@5 & Cov. & Cand.\ Red. \\
\midrule
1 & 34.9\% & 46.4\% & 49.8\% & 96.5\% & 42.6\% & 57.1\% & 65.0\% & 97.2\% \\
3 & 47.1\% & 63.0\% & 67.8\% & 89.8\% & 53.7\% & 71.8\% & 80.5\% & 92.3\% \\
5 & 51.8\% & 69.5\% & 77.3\% & 83.5\% & \textbf{58.2\%} & \textbf{76.9\%} & \textbf{89.2\%} & \textbf{87.4\%} \\
7 & 55.3\% & 73.9\% & 84.4\% & 77.6\% & 60.4\% & 79.2\% & 92.3\% & 82.5\% \\
10 & 59.6\% & 79.0\% & 90.3\% & 69.6\% & 62.1\% & 82.6\% & 95.4\% & 75.2\% \\
\midrule
\textit{Tool-only ref.} & \textbf{66.1\%} & \textbf{87.2\%} & \textbf{100.0\%} & 0.0\% & \multicolumn{4}{c}{\textit{accuracy anchor}} \\
\bottomrule
\end{tabular}
}
\caption{Main StableToolBench hierarchical-routing Pareto table. Tool-only retrieval is the accuracy anchor; the hierarchy rows trade quality for candidate-context reduction.}
\label{tab:hierarchy-main}
\end{table*}

\begin{table*}[t]
\centering
\footnotesize
\setlength{\tabcolsep}{6pt}
\resizebox{.9\textwidth}{!}{%
\begin{tabular}{lrrrrrr}
\toprule
Routing Mode & Cand. Tok. & Final Tok. & Cov. & P@1 & R@5 & Cand.\ Red. \\
\midrule
Full Catalog Injection & 172,426 & 172,426 & 100.0\% & -- & -- & 0.0\% \\
Tool-Only Retrieval & 172,426 & 407 & 100.0\% & 66.1\% & 87.2\% & 0.0\% \\
Manual Skill $\rightarrow$ Tool (\texttt{k=5}) & 28390 & 410 & 77.3\% & 51.8\% & 69.5\% & 83.5\% \\
Automatic Skill $\rightarrow$ Tool (\texttt{k=5}) & \textbf{21785} & 411 & 89.2\% & 58.2\% & 76.9\% & \textbf{87.4\%} \\
\bottomrule
\end{tabular}
}
\caption{Routing-stage token efficiency on StableToolBench. Full catalog versus tool-only explains final-prompt compression, while tool-only versus hierarchy explains candidate-pool compression.}
\label{tab:token-efficiency}
\end{table*}

The benchmark-native local-execution check keeps only the two most policy-relevant hierarchy conditions: the official benchmark-provided exposure baseline and \texttt{auto\_skill\_tool}. Under exact validation against the local StableToolBench tool store, this yields the complete shared executable subset across \emph{all six} StableToolBench groups: \texttt{74} queries in \texttt{G1\_instruction}, \texttt{65} in \texttt{G1\_category}, \texttt{57} in \texttt{G1\_tool}, \texttt{26} in \texttt{G2\_instruction}, \texttt{16} in \texttt{G2\_category}, and \texttt{21} in \texttt{G3\_instruction}, for a total of \texttt{259}. Table~\ref{tab:appendix-stb-exec} reports the resulting benchmark-native execution comparison under the official StableToolBench execution code, a fixed local \texttt{Qwen-14B} planner, and the same local function-call-accuracy-style judge used throughout this appendix. On that larger paired set, \texttt{auto\_skill\_tool} slightly exceeds the official exposure baseline overall (\texttt{32.8\%} vs.\ \texttt{31.7\%}) while also improving token cost per solved task (\texttt{17.7k} vs.\ \texttt{18.3k}). This is the only hierarchy-side execution result we keep in the appendix because it is the one that directly supports the main-text efficiency claim.

\begin{table}[ht]
\centering
\footnotesize
\setlength{\tabcolsep}{4pt}
\begin{tabular}{lrrrr}
\toprule
Group & Count & Official & Auto & $\Delta$ Auto \\
\midrule
\texttt{G1\_instruction} & 74 & 35.1\% & 31.1\% & -4.1 \\
\texttt{G1\_category} & 65 & 33.8\% & 46.2\% & +12.3 \\
\texttt{G1\_tool} & 57 & 31.6\% & 29.8\% & -1.8 \\
\texttt{G2\_instruction} & 26 & 26.9\% & 26.9\% & +0.0 \\
\texttt{G2\_category} & 16 & 25.0\% & 6.2\% & -18.8 \\
\texttt{G3\_instruction} & 21 & 23.8\% & 33.3\% & +9.5 \\
\midrule
\textbf{All 6 groups} & \textbf{259} & \textbf{31.7\%} & \textbf{32.8\%} & \textbf{+1.1} \\
\bottomrule
\end{tabular}
\caption{Benchmark-native local execution on the \texttt{259}-query paired common subset shared by \texttt{official\_api\_list} and \texttt{auto\_skill\_tool} across all six StableToolBench groups. The two methods use the same local \texttt{Qwen-14B} planner and the same local function-call-accuracy-style judge. Average end-to-end token counts stay nearly identical (\texttt{5,807} vs.\ \texttt{5,822}) because planner/execution overhead dominates the full trace length, but token cost per solved task still favors the automatic hierarchy (\texttt{17.7k} vs.\ \texttt{18.3k}).}
\label{tab:appendix-stb-exec}
\end{table}

\end{document}